\documentclass{article} % For LaTeX2e
\usepackage{iclr2025_conference,times}

\usepackage{amsmath,amsfonts,bm}

\def\eqref#1{equation~\ref{#1}}
\def\1{\bm{1}}

\DeclareMathAlphabet{\mathsfit}{\encodingdefault}{\sfdefault}{m}{sl}
\SetMathAlphabet{\mathsfit}{bold}{\encodingdefault}{\sfdefault}{bx}{n}

\usepackage{hyperref}
\usepackage{url}

\usepackage{pifont}
\usepackage{graphicx}
\usepackage{overpic}
\usepackage{booktabs}
\usepackage{makecell}
\usepackage{multirow}
\usepackage{wrapfig}
\usepackage{marvosym}
\usepackage{caption}
\usepackage{colortbl}
\usepackage{xcolor}         % colors

\title{FakeShield: Explainable Image Forgery Detection and Localization via Multi-modal Large Language Models}

\author{Zhipei Xu$^{1,2}$, Xuanyu Zhang$^{1}$, Runyi Li$^{1}$, Zecheng Tang$^{1}$, Qing Huang$^{4}$, Jian Zhang$^{1,2,3 \dagger}$
\\
$^1$School of Electronic and Computer Engineering, Peking University\\
$^2$Peking University Shenzhen Graduate School-Rabbitpre AIGC Joint Research Laboratory\\
$^3$Guangdong Provincial Key Laboratory of Ultra High Definition Immersive Media Technology, \\ Shenzhen Graduate School, Peking University\\
$^4$School of Future Technology, South China University of Technology\\
}

\iclrfinalcopy % Uncomment for camera-ready version, but NOT for submission.
\begin{document}

\let\thefootnote\relax\footnotetext{$\dagger$ Corresponding author: Jian Zhang. This work was supported in part by Guangdong Provincial Key Laboratory of Ultra High Definition Immersive Media Technology (No. 2024B1212010006) and Shenzhen General Research Project (No. JCYJ20241202125904007).}

\maketitle

\begin{abstract}

The rapid development of generative AI is a double-edged sword, which not only facilitates content creation but also makes image manipulation easier and more difficult to detect. Although current image forgery detection and localization (IFDL) methods are generally effective, they tend to face two challenges: \textbf{1)} black-box nature with unknown detection principle, \textbf{2)} limited generalization across diverse tampering methods (e.g., Photoshop, DeepFake, AIGC-Editing). To address these issues, we propose the explainable IFDL task and design FakeShield, a multi-modal framework capable of evaluating image authenticity, generating tampered region masks, and providing a judgment basis based on pixel-level and image-level tampering clues. Additionally, we leverage GPT-4o to enhance existing IFDL datasets, creating the Multi-Modal Tamper Description dataSet (MMTD-Set) for training FakeShield's tampering analysis capabilities. Meanwhile, we incorporate a Domain Tag-guided Explainable Forgery Detection Module (DTE-FDM) and a Multi-modal Forgery Localization Module (MFLM) to address various types of tamper detection interpretation and achieve forgery localization guided by detailed textual descriptions. Extensive experiments demonstrate that FakeShield effectively detects and localizes various tampering techniques, offering an explainable and superior solution compared to previous IFDL methods. The code is available at \url{https://github.com/zhipeixu/FakeShield}.
\end{abstract}

% 他的黑盒性质（检测原理未知）

\section{Introduction}

% With the rapid development of AIGC, powerful image generation models have provided a breeding ground for convenient image tampering, blurring the boundaries between true and forgery.

% Thanks to the rapid development of diffusion models, the vigorous growth of AI-generated models has enabled effective image editing methods such as DALL·E 3 [19], Imagen [64], and Stable Diffusion [63] to generate and edit highly realistic images. However, this is a double-edged sword. On the one hand, their emergence greatly facilitates the work of photographers, illustrators, and others, allowing them to efficiently process and generate images. On the other hand, it has also led to an increase in malicIoUs tampering and illegal appropriation. The fake images altered by AIGC technology are difficult to distinguish from real images with the naked eye, making it difficult to ensure the authenticity of images on social media, which will lead to problems such as social unrest, economic losses, and legal concerns.

With the rapid development of AIGC, powerful image editing models have provided a breeding ground for convenient image tampering, blurring the boundaries between true and forgery. People can use AIGC image editing methods~\citep{rombach2022high,zhang2023adding,suvorov2022resolution,mou2023dragondiffusion} to edit images without leaving a trace. Although it has facilitated the work of photographers and illustrators, AIGC editing methods have also led to an increase in malicious tampering and illegal theft. The authenticity of images in social media is difficult to guarantee, which will lead to problems such as rumor storms, economic losses, and legal concerns.
Therefore, it is important and urgent to identify the authenticity of images.
In this context, the image forgery detection and localization (IFDL) task aims to identify whether an image has been tampered with and locate the specific manipulation areas. It can be widely applied in the real world, such as filtering false content on social media, preventing
the spread of fake news, and court evidence collection.

\begin{figure}[t]
	\centering
	\includegraphics[width=0.95\linewidth]{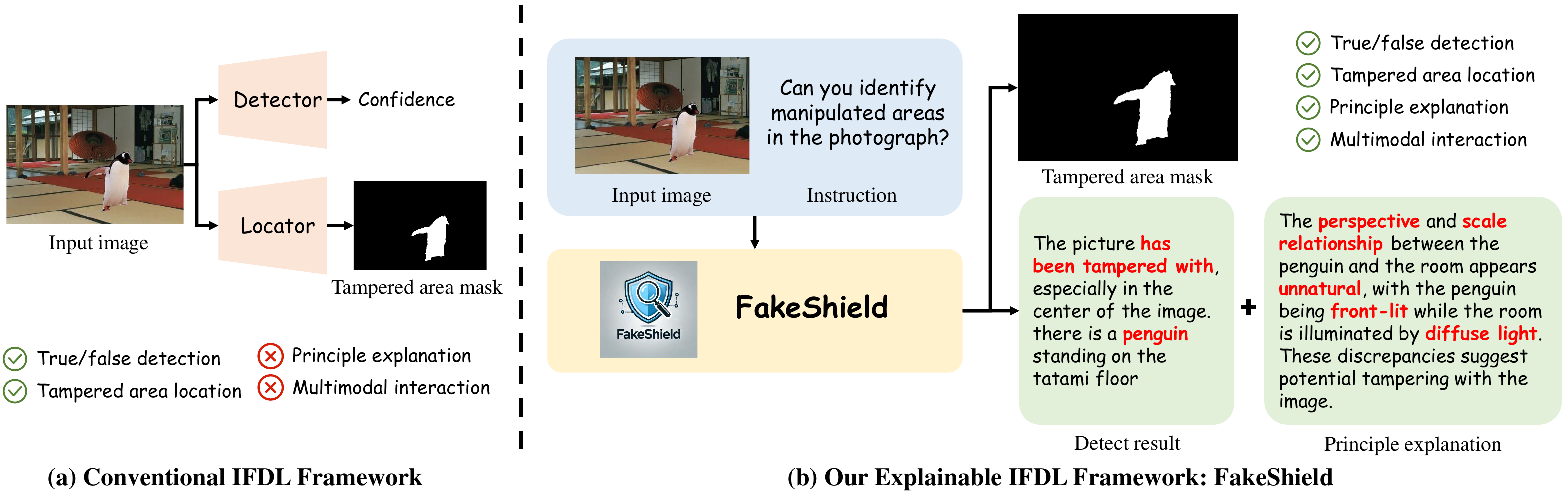}
	\vspace{-5pt}
	\caption{Illustration of the conventional IFDL and explainable IFDL framework. Conventional methods offer only detection results and tampered masks. We extend this into a multi-modal framework, enabling detailed explanations and conversational interactions for a deeper analysis.}
	\label{teasor}
    \vspace{-15pt}
\end{figure}

State-of-the-art IFDL methods have utilized well-designed network structures, elaborate network constraints, and efficient pre-training strategies to achieve remarkable performance~\citep{yu2024diffforensics,ma2023iml,dong2022mvss}.
However, previous IFDL methods face two key problems, limiting their practicality and generalizability. \textcolor{blue}{\textbf{First}}, as shown in Figure~\ref{teasor}(a), most existing IFDL methods are black-box models, only providing the authenticity probability of the image, while the principle of detection is unknown to users. Since the existing IFDL methods cannot guarantee satisfactory accuracy, manual subsequent judgment is still required. Given that the information provided by the IFDL methods is insufficient, it is difficult to support the human assessment and users still need to re-analyze the suspect image by themselves.
\textcolor{blue}{\textbf{Second}}, 
in real-world scenarios, tampering types are highly diverse, including Photoshop (copy-and-move, splicing, and removal), AIGC-Editing, DeepFake, and so on. Existing IFDL methods~\citep{yu2024diffforensics,ma2023iml} are typically limited to handling only one of these techniques, lacking the ability to achieve comprehensive generalization. This forces users to identify different tampering types in advance and apply specific detection methods accordingly, significantly reducing these models' practical utility.

% 现实生活中的篡改类型非常广泛，PhotoShop，AIGC-Editing，DeepFake等各种篡改方法层出不穷，现有的IFDL方法只能处理其中的一到两类，无法实现真正的全面的泛化。这就要求用户事先区分不同的篡改类型，使用不同的检测方法，这大大限制了模型的实用价值。

%  Other methods) also, adopt pre-trained diffusion denoisers to effectively capture local tampered cues

% 传统的IFDL方法仅提供检测结果和篡改区域mask，无法解释判断原理，也没有任何形式的交互，让用户进一步理解被伪造的图片
% eIFDL方法，还提供了对检测结果的解释，并且可以提供多模态的对话式交互

Benefiting from the rapid advancements in Transformer architectures, Large Language Models (LLMs) have attracted significant attention. Furthermore,~\citep{liu2024visual} introduced a Multi-modal Large Language Model (M-LLM) that aligns visual and textual features, thereby endowing LLMs with enhanced visual comprehension abilities. Given that LLMs are pre-trained on an extensive and diverse corpus of world knowledge, they hold significant potential for a wide range of applications, such as machine translation~\citep{devlin2018bert}, code completion, and visual understanding~\citep{liu2024visual}. Consequently, we explored the feasibility of employing M-LLMs for explainable Image Forgery Detection and Localization (e-IFDL). This approach allows for a more comprehensive explanation of the rationale behind tampering detection and provides a more precise identification of both the authenticity of images and the suspected manipulation regions.

To address the two issues of the existing IFDL methods, we propose the explainable-IFDL (e-IFDL) task and a multi-modal explainable tamper detection framework called FakeShield. As illustrated in Figure\ref{teasor}(b), the e-IFDL task requires the model to evaluate the authenticity of any given image, generate a mask for the suspected tampered regions, and provide a rationale based on some pixel-level artifact details (e.g., object edges, resolution consistency) and image-level semantic-related errors (e.g., physical laws, perspective relationships).
Leveraging the capabilities of GPT-4o~\citep{openai2023gpt}, we can generate a comprehensive triplet consisting of a tampered image, a modified area mask, and a detailed description of the edited region through a meticulously crafted prompt. 
Then, we develop the \textbf{M}ulti-Modal \textbf{T}amper \textbf{D}escription data\textbf{Set} (MMTD-Set) by building upon existing IFDL datasets. Utilizing the MMTD-Set, we fine-tune M-LLM~\citep{liu2024visual} and visual segmentation models~\citep{kirillov2023segment,lai2024lisa}, equipping them with the capability to provide complete analysis for judgment, detecting tampering, and generate accurate tamper area masks. This process ultimately forms a comprehensive forensic pipeline for analysis, detection, and localization. Our contributions are summarized as follows:

\vspace{1pt}
\noindent \ding{113}~(1) We present the first attempt to propose a multi-modal large image forgery detection and localization model, dubbed \textbf{FakeShield}. It can not only decouple the detection and localization process but also provide a reasonable judgment basis, which alleviates the black-box property and unexplainable issue of existing IFDL methods.

\vspace{1pt}
\noindent \ding{113}~(2) We use GPT-4o to enrich the existing IFDL dataset with textual information, constructing the \textbf{MMTD-Set}. By guiding it to focus on distinct features for various types of tampered data, GPT-4o can analyze the characteristics of tampered images and construct "image-mask-description" triplets.

\vspace{1pt}
\noindent \ding{113}~(3) We develop a \textbf{D}omain \textbf{T}ag-guided \textbf{E}xplainable \textbf{F}orgery \textbf{D}etection \textbf{M}odule (\textbf{DTE-FDM}) to spot different types of fake images in a united model and effectively alleviate the data domain conflict. Meanwhile, an \textbf{M}ulti-modal \textbf{F}orgery \textbf{L}ocalization \textbf{M}odule (\textbf{MFLM}) is adopted to align visual-language features, thus pinpointing tampered areas.

\vspace{1pt}
\noindent \ding{113}~(4) Extensive experiments demonstrate that our method can accurately analyze tampering clues, and surpass most previous IFDL methods in the detection and localization of many tampering types like copy-move, splicing, removal, DeepFake, and AIGC-based editing.
\vspace{-10pt}

\section{Related works}
\vspace{-10pt}

\subsection{Image Forgery Detection and Localization}
\vspace{-10pt}

Prevailing IFDL methods mainly target at the localization of specific manipulation types~\citep{salloum2018image, islam2020doa, li2018fast, zhu2018deep, li2019localization}. In contrast, universal tamper localization methods~\citep{li2018learning, kwon2021cat, chen2021image, ying2023learning, ying2021image, hu2023draw, ying2022rwn, li2024protect, yu2024cross, zhang2024gshider} aim to detect artifacts and irregularities across a broader spectrum of tampered images. For instance, MVSS-Net~\citep{dong2022mvss} utilized multi-scale supervision and multi-view feature learning to simultaneously capture image noise and boundary artifacts. OSN~\citep{wu2022robust} employed a robust training strategy to overcome the difficulties associated with lossy image processing. HiFi-Net~\citep{guo2023hierarchical} adopted a combination of multi-branch feature extraction and localization modules to effectively address alterations in images synthesized and edited by CNNs.
IML-ViT~\citep{ma2023iml} integrated Swin-ViT into the IFDL task, employing an FPN architecture and edge loss constraints to enhance its performance. DiffForensics~\citep{yu2024diffforensics} adopted a training approach akin to diffusion models, strengthening the model's capacity to capture fine image details. 
Additionally, some researchers~\citep{zhang2024editguard,zhang2024v2a,asnani2023malp} have pursued proactive tamper detection and localization by embedding copyright and location watermarks into images/audio/videos preemptively. However, despite their acceptable performances, these IFDL methods cannot explain the underlying principles and rationale behind their detection and localization judgments, offering no interaction. Moreover, they suffer from limited generalization and accuracy, exhibiting significant performance disparities across different testing data domains.
\vspace{-5pt}

\subsection{Large Language Model}
\vspace{-5pt}

% 近年来，大语言模型由于其卓越的指令跟随和文本生成能力，取得了举世瞩目。Built on the Transformer architecture, these models are pre-trained on vast amounts of text data, enabling them to acquire extensive world knowledge that aids in the generalization of downstream tasks.
% 接着，一些学者通过加入图像编码器和投影层，将图片编码为与文本对齐的token，从而将LLM强大的理解能力和世界知识扩展到视觉模态。
% 这些工作通过扩展视觉指令微调数据集和模型规模，赋予了 LVLM 强大的视觉理解能力。

% 目前，M-LLM在许多下游任务中也有非常瞩目的表现。LISA通过将SAM与M-LLM结合，实现了reason segment，可以实现从文本到分割mask。Glamm使用更复杂和先进的图像编码器，更进一步的实现了文本和mask的grounding。另外，也有部分工作探索了多模态大语言模型在图像篡改检测领域的表现，~\citep{yang2024research}使用qwen，gpt等预训练M-LLM，测试其在IML和DeepFake数据集上的性能，但结果较差。最近，张等人通过手动结合对某些可以根据常识辨别的真实和伪造面孔的评级推理过程来构建 DD-VQA 数据集，微调M-LLM可以实现DeepFake检测，但其仅适用于DeepFake数据域的图像篡改检测，且其推理过程简短，不够详细

% 综上，现有的大语言模型都无法实现对于篡改图片的检测，定位和解释，

Large language models~\citep{dubey2024llama,openai2023gpt} have garnered global attention in recent years for their exceptional instruction-following and text-generation abilities. 
Based on the Transformer architecture, LLMs are pre-trained on massive datasets, allowing them to accumulate broad world knowledge that enhances their ability to generalize across a wide range of downstream tasks. 
Subsequently, some researchers~\citep{li2022blip} expanded LLMs' powerful understanding and world knowledge to the visual domain by incorporating image encoders and projection layers, which enable images encoded into tokens that align with the text. 
Some recent works~\citep{chen2023sharegpt4v,wang2023cogvlm,chen2023bianque} equipped M-LLMs with enhanced visual understanding capabilities by expanding the visual instruction datasets and increasing the model size during fine-tuning. 
Currently, M-LLMs demonstrate impressive performance across various downstream tasks. LISA~\citep{lai2024lisa} integrated SAM~\citep{kirillov2023segment} with M-LLM to implement reasoning segmentation, enabling the generation of masks from text descriptions. GLaMM~\citep{rasheed2024glamm} further enhanced this by using a more advanced region image encoder to improve text-to-mask grounding. Additionally, some studies~\citep{yang2024research,zhang2024common} have explored the application of M-LLMs in DeepFake detection. For instance,~\citep{zhang2024common} introduced the DD-VQA dataset, combining a manual inference process for rating real and fake faces that can be distinguished using common sense. 
Targeted at Deepfake detection, \citep{huang2024ffaa} used GPT-4o to create image-analysis pairs, and introduced a multi-answer intelligent decision system into MLLM, achieving good effect. However, it cannot be generalized to other types of tampering such as Photoshop and AIGC-Editing, and cannot accurately locate the tampered areas. Besides, using M-LLMs to realize universal tamper localization and detection remains unexplored.

% However, using M-LLMs to realize universal tamper localization and detection remains unexplored.

\section{Methodology}
\vspace{-5pt}
% Image $\mathbf{I}_{ori}$, Text $\mathbf{T}_{ins}$, $\mathbf{T}_{tag}$, Interpretation $\mathbf{T}_{det}$,
% Output
% $\mathbf{O}_{det}$,
% Mask
% $\mathbf{M}_{loc}$

% 总述

% 在这一部分，我们首先介绍为了训练介绍强大的FakeShield框架，其包括

% This section introduces the Multi-Modal Tamper Description Dataset (MMTD-Set) and FakeShield. MMTD-Set uses GPT-4o to generate detailed text descriptions for tampered images. FakeShield includes a Domain Tag-guided Detection and a Multi-modal Localization Module, improving tampering detection and localization with specialized prompts and data.

\subsection{Construction of the proposed MMTD-Set}
\label{Construction of the proposed MMTD-Set}

% 构建数据集，pipeline画个图，给点例子，
% 1. 动机：现有数据集缺乏多模态数据，缺少对于篡改区域的文字描述分析
% 2. 构建数据集的原则，详细展开 pixel和image
% 3. 给一些Prompt模版，然后给一些例子，引用图片
% 4. 描述一下实现细节，如何使用gpt生成的，引用一下图片

% 为什么提：缺乏适配于训练多模态大模型的训练集
% 困难之处：如何高效的批量的将视觉表示的篡改信息转化为精确的文本描述。从而支撑多模态大模型的训练

% 其构建的困难之处，在于如何将现有IFDL数据集的视觉表示篡改信息，转化为精确的文本描述

% 重要性：
% 把mask给gpt，让其准确感知在哪篡改
% 对于篡改依据的设计，引导。
% 更进一步，我们对数据集的真假图的Prompt设计做了如下的设计

% 目前，由于缺少适配于训练多模态大模型的数据集，
% 为了解决这一问题，我们借助GPT，同时把篡改图片和mask提供给它，让其准确感知篡改位置。并巧妙设计指令，引导其输出合适的描述内容。更进一步，我们对于数据集中的真图和假图做了不同的指令设计：

% given一张待检测的图片，他被输入给Domain tag generator，得到一个数据域tag。将tag，文本query和img token同时输入给微调后的LLM，可以得到篡改检测结果与说明O。随后将O与原图片I一起输入给Tamper C Module，其结果作为SAM分割的Prompt，引导SAM对于原图片mask的生成

% 对于真实图像，我们仍然要求其从相同的角度分析，如何确认图像的真实性，将其与篡改图片区分开来

% Note that during the dataset construction process, GPT is directly informed of the image's authenticity through the instructions and is required to describe the image from the perspective of a human detector, without explicitly mentioning the mask-related information provided.

% 用tamper-type-specific prompts引导gpt4o分析图片，构建judgment basis，关注pixel-level details and image-level content

\begin{figure}[t!]
	\centering
	\includegraphics[width=0.9\linewidth]{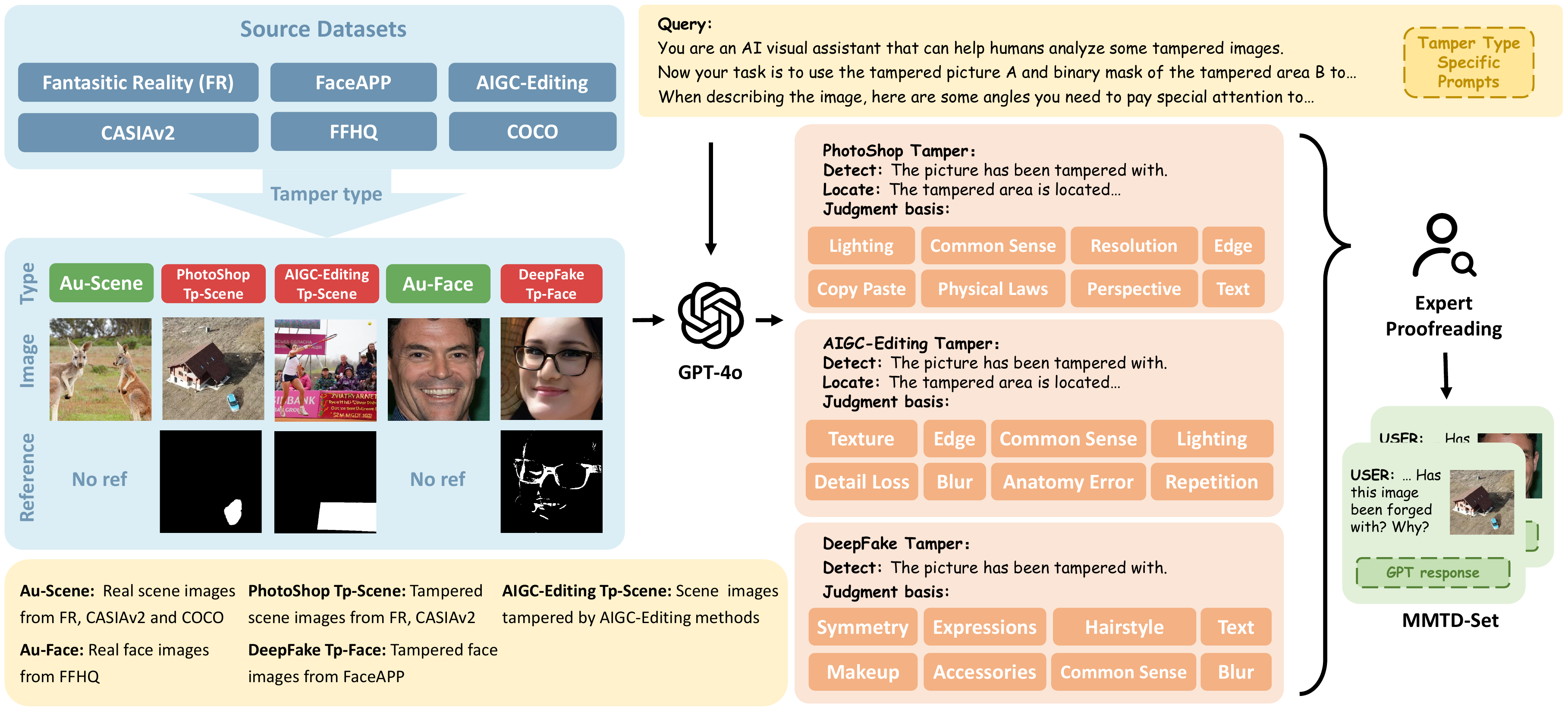}
	\vspace{-5pt}
	\caption{Illustration of the construction process of our MMTD-Set. We sample the tampered image-mask pairs from PS, DeepFake, and AIGC benchmarks, and then use domain tags to guide GPT-4o in constructing the judgment basis and focusing on both pixel-level details and image-level content.}
	\label{dataset}
    \vspace{-16pt}
\end{figure}

% 数据收集，数据来源，数据分类，结合图的左半边说，说一下一共多少数据量
% GPT辅助生成：因为人工描述耗时耗力，受LLaVA启发，使用gpt辅助生成分析和描述非常的好用，如图，描述一下这个图。

% 首先，我们将常见的图片篡改类型分为三个类别：Photoshop篡改：包括splicing，copymove，removal等常见的传统篡改操作；DeepFake篡改：主要包括使用faceapp等面部属性篡改方法，局部篡改方式，

\textbf{Motivation:} Most existing IFDL datasets consist of a single visual modality, lacking training visual-language samples adapted to M-LLMs. The challenge of constructing our MMTD-Set lies in accurately translating the visual tampering information from the existing IFDL image datasets into precise textual descriptions. To address this challenge, our core contributions focus on two aspects: \textbf{(1)} We leverage GPT-4o to generate text description and provide both the tampered image and its corresponding mask to GPT-4o, enabling it to accurately identify the tampered location. \textbf{(2)} For each tamper type, we design specific prompts to their unique characteristics, guiding GPT-4o to focus on different tampering artifacts and providing more detailed visual cues.

\textbf{Data collection:} Based on~\citep{ma2023iml,nirkin2021DeepFake}, we categorize common tampering into three types: PhotoShop (copy-move, splicing, removal), DeepFake (FaceAPP~\citep{faceapp2017}), and AIGC-Editing (SD-inpainting~\citep{Lugmayr_2022_CVPR}). As shown in Figure~\ref{dataset}, we gathered three types of tampered images along with their corresponding authentic images from public datasets~\citep{dong2013casia,dang2020detectiondffd} and self-constructed data.

\textbf{GPT assisted description generation: }Given that manual analysis of tampered images is time-consuming, inspired by~\citep{liu2024visual,chen2023sharegpt4v,huang2024ffaa}, we used GPT-4o to automate the analysis of tampered images. As depicted in Figure~\ref{dataset}, the output analysis is required to follow the format of detected results, localization descriptions, and judgment basis.

\textbf{For tampered images}, we input the edited image, its corresponding forgery mask, and our carefully constructed tamper type specific prompts into the powerful GPT-4o to more accurately describe the tampered regions. \textbf{For authentic images}, GPT-4o is provided with only the real image and a set of prompts, guiding it to confirm its authenticity. \textit{The full-text prompts are detailed in the Appendix~\ref{Appendix Prompts}.} To more clearly and specifically describe and analyze the tampering of images, GPT-4o describes the image from two key aspects: the location and content of the tampered areas, and any visible artifacts or semantic errors caused by the tampering: \textbf{(1)} For the tampering location, GPT-4o is required to describe it in both absolute positions (e.g., top, bottom, upper left corner, lower right corner) and relative positions (e.g., above the crowd, on the table, under the tree). When analyzing the tampered content, it is tasked with providing detailed information about the types, quantities, actions, and attributes of the objects within the tampered region.
\textbf{(2)} For the visible artifacts and semantic errors, since different tampering methods produce distinct types of artifacts, we craft specific prompts to guide the analysis. It can broadly be categorized into pixel-level artifact details and image-level semantic-related errors. For PhotoShop (PS) tampering, operations like copy-move and splicing often introduce pixel-level issues such as edge artifacts, abnormal resolution, and inconsistencies in lighting. Additionally, semantic-level errors, including violations of physical laws or common sense, are frequently observed. In AIGC-Editing (AIGC), for instance, it often fails to generate text accurately, resulting in disordered symbols or characters appearing in the tampered area. For the DeepFake (DF), tampering with facial features frequently results in localized blurring. 
% Further details are illustrated in Figure~\ref{dataset} and Appendix~\ref{Appendix Examples}.

% \textbf{For authentic images}, to balance the positive and negative samples, authentic images are also included in the proposed MMTD-Set. As shown in Figure~\ref{dataset}, GPT-4o is provided with only the real image and a set of prompts, guiding it to analyze the image from the same perspective as mentioned earlier, with a focus on confirming its authenticity and distinguishing it from tampered ones.

\begin{figure}[t]
	\centering
	\includegraphics[width=1\linewidth]{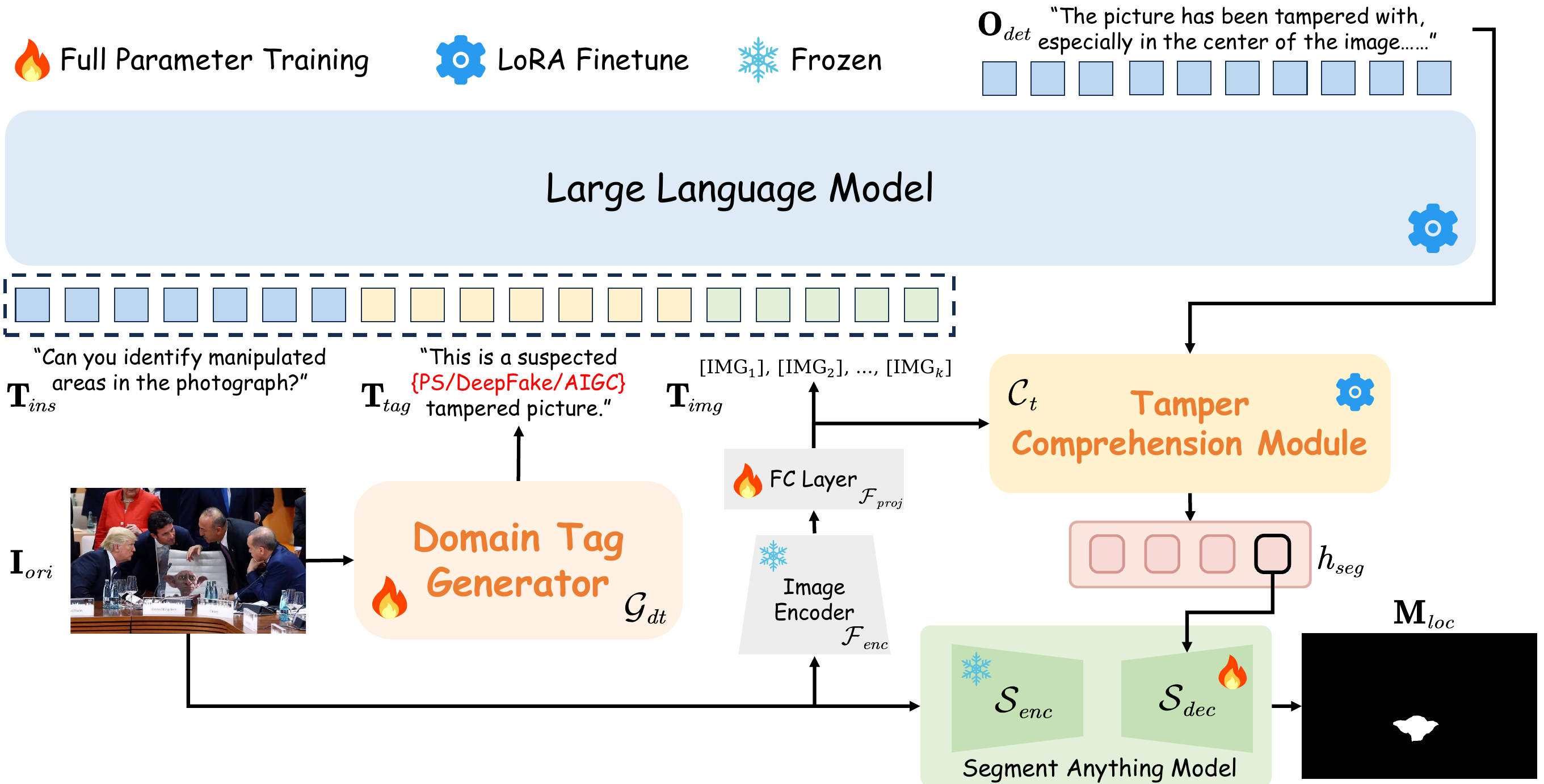}
	% \vspace{-18pt}
	\caption{The pipeline of FakeShield. Given an image $\mathbf{I}_{ori}$ for detection, it is first processed by the Domain Tag Generator $\mathcal{G}_{dt}$ to obtain a data domain tag $\mathbf{T}_{tag}$. The tag $\mathbf{T}_{tag}$, along with the text instruction $\mathbf{T}_{ins}$ and image tokens $\mathbf{T}_{img}$, are simultaneously input into the fine-tuned LLM, generating tamper detection result and explanation $\mathbf{O}_{det}$. Subsequently, $\mathbf{O}_{det}$ and $\mathbf{T}_{img}$ are input into the Tamper Comprehension Module $\mathcal{C}_{t}$, and the last-layer embedding for the \texttt{<SEG>} token $\mathbf{h}_{\texttt{<SEG>}}$ serves as a prompt for SAM, guiding it to generate the tamper area mask $\mathbf{M}_{loc}$.}
	\label{pipeline}
    \vspace{-10pt}
\end{figure}

\subsection{Overall Framework of FakeShield}
Our goals involve two issues: \textbf{1):} Utilizing the textual understanding ability and world knowledge of the M-LLM to analyze and judge the authenticity of images; \textbf{2):} Adopting the analysis and interpretation of tampered images to assist the segmentation model in pinpointing the tampered areas. To solve these two tasks, an intuitive approach is to fine-tune a large multimodal model to simultaneously output analysis and tampered masks. However, we find that joint training of multiple tasks will increase the difficulty of network optimization and interfere with each other. Considering that detection and interpretation focus more on language understanding and organization, while localization requires more accumulation of visual prior information, the proposed FakeShield contains two key decoupled parts, namely DTE-FDM and MFLM, as illustrated in Fig.~\ref{pipeline}. Specifically, an original suspected image $\mathbf{I}_{ori}$ and an instruction text $\mathbf{T}_{ins}$ (e.g. ``Can you identify manipulated areas in the photograph?'') are fed to the proposed DTE-FDM to predict the detection result and judgment basis $\mathbf{O}_{det}$. In this process, we use a learnable generator to produce a domain tag $\mathbf{T}_{tag}$, thus avoiding the tampered data domain conflict. Furthermore, we input the interpretation $\mathbf{O}_{det}$ and the image $\mathbf{I}_{ori}$ to the MFLM to accurately extract the tampered mask $\mathbf{M}_{loc}$. To promote cross-modal interaction for tamper localization, we introduce a tamper comprehension module to align the visual and textual features and enhance the ability of the vision foundation model to understand long descriptions.

\subsection{Domain Tag-guided Explainable Forgery Detection Module}

% domain-tag的动机，问题写一段。
%输入输出的piepline写一段。Domain-tag怎么做的
% 公式 1. LLaVA 公式组

% As shown in Table.~\ref{}, various IFDL datasets use different image forgery methods, such as copy-move, splicing, DeepFake, and AIGC-based. However, due to significant differences in data domains, few existing IFDL methods can adapt well to all image forgery methods. To address the data domain discrepancies, drawing inspiration from~\citep{sanh2022multitask}, we introduced the \textbf{Domain Tag-guided Explainable Forgery Detection Module (DTE-FDM)}. This module leverages a specialized prompt, referred to as a domain tag, to guide the model in distinguishing between different data domains.

% 现实生活中，图片可能遭受各种各样方法的篡改和攻击，包括copy-move, splicing, DeepFake, and AIGC-based methods.
\textbf{Motivation:} In real-life scenarios, images can be tampered with and attacked through various methods, including copy-move, splicing, removal, DeepFake, and AIGC-based methods.
However, these tampered images have different distribution characteristics, and domain differences, making it difficult to apply a single IFDL method to all forgery data. 
% 例如，DeepFake关注于人脸修改，其篡改常常会导致面部的局部模糊，以及嘴唇，牙齿和眼部的不自然。然而，像PhotoShop（splicing，copymove）之类的，其往往会在篡改区域的边缘留下较明显的伪影。对于AIGC-Editing操作，他的特点在于篡改区域的模糊，往往会确实一些纹理信息
For example, DeepFake focuses on face modification, often causing partial blurring and unnatural features in the lips, teeth, and eyes. In contrast, tools like PhotoShop (splicing, copy-move, removal) tend to leave noticeable artifacts at the edges of the tampered areas. In the case of AIGC-Editing, blurring within the tampered region often alters or obscures texture details.
To mitigate these significant domain discrepancies, inspired by~\citep{sanh2022multitask}, we introduce the \textbf{Domain Tag Generator(DTG)}, which utilizes a specialized domain tag to prompt the model to distinguish between various data domains.

First, the original image $\mathbf{I}_{ori}$ is input into a classifier $\mathcal{G}_{dt}$ to obtain the domain tag $\mathbf{T}_{tag}$. Specifically, we classify all common tampering types into three categories: Photoshop-based editing, DeepFake, and AIGC-based tampering, and use the template \textit{``This is a suspected \{data domain\}-tampered picture."} as the identifier. 
Simultaneously, consistent with~\citep{liu2024visual}, $\mathbf{I}_{ori}$ is passed through the image encoder $\mathcal{F}_{enc}$ and linear projection layer $\mathcal{F}_{proj}$ to generate the image tokens [\texttt{IMG}] $\mathbf{T}_{img}$. 
Next, $\mathbf{T}_{tag}$ and $\mathbf{T}_{img}$ are concatenated with the instruction $\mathbf{T}_{ins}$ and then fed into the $\operatorname{LLM}$. To be noted, $\mathbf{T}_{ins}$ is a prompt that instructs the model to detect tampering and describe the location of the manipulation, for example: \textit{``Can you identify manipulated areas in the photograph?"}. After several autoregressive predictions, the output $\mathbf{O}_{det}$ comprises three components: detection results, a description of the location of tampered area, and the interpretive basis for the detection. 
\begin{equation}
    \mathbf{T}_{tag} = \mathcal{G}_{dt}(\mathbf{I}_{ori}), \quad \mathbf{T}_{img} = \mathcal{F}_{proj}(\mathcal{F}_{enc}(\mathbf{I}_{ori}))
\end{equation}
\begin{equation}
    \mathbf{O}_{det} = \operatorname{LLM}(\mathbf{T}_{ins}, \mathbf{T}_{tag}  \mid \mathbf{T}_{img} ).     
\end{equation}

Given the large size of LLMs and limited computational resources, full parameter training is impractical. Thus, we freeze the LLM and leverage LoRA fine-tuning technology~\citep{hu2022lora} to preserve semantic integrity while enabling efficient image forgery detection.
% \begin{equation}
%     \hat {\mathbf{W}}=\mathbf{W}+\Delta \mathbf{W}=\mathbf{W}+\mathbf{W}_{down}\mathbf{W}_{up},
% \end{equation}

% where $\mathbf{W}_{down}\in\mathbb{R}^{r\times k}$ and $\mathbf{W}_{up}\in\mathbb{R}^{d\times r}$ are the weight matrices of the LoRA, $r$ is much smaller than $d$ and $k$. $\mathbf{W}_{}$ is initialized to zero to ensure that LoRA does not corrupt the original output.

\subsection{Multi-modal Forgery Localization Module}

\textbf{Motivation:} Although $\mathbf{O}_{det}$ provides a textual description of the tampered area, it lacks precision and intuitive clarity. To address this issue, we aim to transform $\mathbf{O}_{det}$ into an accurate binary mask, providing a clearer and more accurate representation of the tampered region. Existing prompt-guided segmentation algorithms~\citep{kirillov2023segment,lai2024lisa} struggle to capture the semantics of long texts and hard to accurately delineate modified regions based on detailed descriptions. Inspired by~\citep{lai2024lisa}, we propose a \textbf{Tamper Comprehension Module (TCM)}, which is an LLM serving as an encoder aligns long-text features with visual modalities, enhancing SAM's precision in locating the forgery areas. To generate the prompt fed into SAM, following~\citep{lai2024lisa}, we introduce a specialized token \texttt{<SEG>}.

As shown in Fig.\ref{pipeline}, the tokenized image $\mathbf{T}_{img}$ and the tampered description $\mathbf{O}_{det}$ are fed into the TCM $\mathcal{C}_{t}$. Then, we extract the last-layer embedding of TCM and transform it into $\mathbf{h}_{\texttt{<SEG>}}$ via an MLP projection layer. 
Simultaneously, the original image $\mathbf{I}_{ori}$ is processed through the SAM encoder $\mathcal{S}_{enc}$ and decoder $\mathcal{S}_{dec}$, where $\mathbf{h}_{\texttt{<SEG>}}$ serve as a prompt for $\mathcal{S}_{dec}$ guiding the mask generation $\mathbf{M}_{loc}$.
\begin{equation}
\centering
\begin{aligned}
    \mathbf{E}_{mid} = \mathcal{S}_{enc}(\mathbf{I}_{ori}), \quad &\mathbf{h}_{\texttt{<SEG>}} = \operatorname{Extract}(\mathcal{C}_{t}(\mathbf{T}_{img},\mathbf{O}_{det})) \\
    \mathbf{M}_{loc} = \mathcal{S}_{dec}&(\mathbf{E}_{mid} \mid \mathbf{h}_{\texttt{<SEG>}}),
\end{aligned}
\end{equation}
where $\mathbf{E}_{mid}$ represents the intermediate features of SAM, and $\operatorname{Extract}(\cdot)$ denotes the operation of extracting the last-layer embedding corresponding to the \texttt{<SEG>} token. Similar to DTE-FDM, we also apply LoRA fine-tuning to MFLM for greater efficiency. With the integration of TCM, SAM will achieve more precise localization of the forgery areas.

\subsection{Training Objectives}
The two submodules of our FakeShield are trained end-to-end separately. \textbf{For DTE-FDM}, the domain tag generator utilizes cross-entropy loss $\ell_{ce}$ as its training objective, enabling it to distinguish between different data domains. Following the approach of LLaVA, our LLM's training objective is the cross-entropy loss $\ell_{ce}$. The training target of DTE-FDM $\ell_{det}$ can be formulated as:
\begin{equation}
    \ell_{det} = \ell_{ce}(\hat{\mathbf{O}}_{det}, \mathbf{O}_{det}) + \lambda \cdot \ell_{ce}(\hat{\mathbf{T}}_{tag}, \mathbf{T}_{tag}),
\end{equation}
% O_det和T_tag分别代表LLM和DTG的预测值，O_det_hat和T_tag_hat分别代表他们对应的真值
where $\lambda$ denotes the weight balancing different loss components, $\mathbf{O}_{det}$ and $\mathbf{T}_{tag}$ represent the predictions of LLM and DTG, while $\hat{\mathbf{O}}_{det}$ and $\hat{\mathbf{T}}_{tag}$ represent their corresponding ground truth. \textbf{For MFLM}, we apply $\ell_{ce}$ to constrain TCM to produce high-quality prompt $\mathbf{y}_{txt}$ with \texttt{<SEG>} token. Meanwhile, we use a linear combination of binary cross-entropy loss $\ell_{bce}$ and dice loss $\ell_{dice}$ to encourage the output of MFLM $\mathbf{M}_{loc}$ to be close to the GT mask $\hat{\mathbf{M}}_{loc}$. Given the ground-truth prompt $\hat{\mathbf{y}}_{txt}$ (e.g., ``It is  \texttt{<SEG>}'') and mask $\hat{\mathbf{M}}_{loc}$, our training losses for MFLM $\ell_{loc}$ can be formulated as:
\begin{equation}
% \begin{aligned}
    \ell_{loc} = \ell_{ce}(\hat{\mathbf{y}}_{txt}, \mathbf{y}_{txt}) + \alpha \cdot \ell_{bce}(\hat{\mathbf{M}}_{loc}, \mathbf{M}_{loc}) + \beta \cdot \ell_{dice}(\hat{\mathbf{M}}_{loc}, \mathbf{M}_{loc}),
% \end{aligned}
\end{equation}

where $\alpha$ and $\beta$ are weighting factors used to balance the respective losses. $\ell_{ce}$, $ \ell_{bce}$, and $\ell_{dice}$ refer to cross-entropy loss, binary cross-entropy loss, and dice loss~\citep{sudre2017generalised} respectively.

\begin{table}[t!]
\centering
\caption{Detection performance comparison between our FakeShield and other competitive methods. Our method can achieve the best detection accuracy in PhotoShop, DeepFake, and AIGC-Editing tampered datasets. The best score is highlighted in \textbf{bold} and the second-best score is \underline{underlined}.}
\vspace{-10pt}
\renewcommand{\arraystretch}{1.3}
\resizebox{1.\linewidth}{!}{
\begin{tabular}{c|cccccccccc|cc|cc}
\toprule[1.5pt]
\multirow{3}{*}{Method} & \multicolumn{10}{c|}{PhotoShop} &  \multicolumn{2}{c|}{\multirow{2}{*}{DeepFake}} & \multicolumn{2}{c}{\multirow{2}{*}{AIGC-Editing}} \\
% \cmidrule(r){2-11}
& \multicolumn{2}{c}{CASIA1+}  & \multicolumn{2}{c}{IMD2020} & \multicolumn{2}{c}{Columbia} & \multicolumn{2}{c}{Coverage} & \multicolumn{2}{c|}{DSO} &\multicolumn{2}{c|}{} \\
\cmidrule(r){2-3} \cmidrule(r){4-5} \cmidrule(r){6-7} \cmidrule(r){8-9} \cmidrule(r){10-11} \cmidrule(r){12-13}  \cmidrule(r){14-15}
& ACC & F1 & ACC & F1 & ACC & F1 & ACC & F1 & ACC & F1 & ACC & F1 & ACC & F1 \\
\hline
SPAN & 0.60 & 0.44 & 0.70 & 0.81 & 0.87 & 0.93 & 0.24 & 0.39 & 0.35 & 0.52 & 0.78 & 0.78 & 0.47 & 0.05   \\
ManTraNet & 0.52 & 0.68 & \underline{0.75} & \underline{0.85} & \underline{0.95} & \underline{0.97} & \underline{0.95} & \underline{0.97} & 0.90 & 0.95 & 0.50 & 0.67 & 0.50 & 0.67  \\
HiFi-Net & 0.46 & 0.44 & 0.62 & 0.75 & 0.68 & 0.81 & 0.34 & 0.51 & \underline{0.96} & \underline{0.98} & 0.56 & 0.61 & 0.49 & 0.42  \\
PSCC-Net & \underline{0.90} & \underline{0.89} & 0.67 & 0.78 & 0.78 & 0.87 & 0.84 & 0.91 & 0.66 & 0.80 & 0.48 & 0.58 & 0.49 & 0.65  \\
CAT-Net & 0.88 & 0.87 & 0.68 & 0.79 & 0.89 & 0.94 & 0.23 & 0.37 & 0.86 & 0.92 & \underline{0.85} & 0.84 & \underline{0.82} & \underline{0.81}  \\
MVSS-Net & 0.62 & 0.76 & \underline{0.75} & \underline{0.85} & 0.94 & \underline{0.97} & 0.65 & 0.79 & \underline{0.96} & \underline{0.98} & 0.84 & \underline{0.91} & 0.44 & 0.24  \\
FakeShield & \textbf{0.95} & \textbf{0.95} & \textbf{0.83} & \textbf{0.90} & \textbf{0.98} & \textbf{0.99} & \textbf{0.97} & \textbf{0.98} & \textbf{0.97} & \textbf{0.98} & \textbf{0.93} & \textbf{0.93} & \textbf{0.98} & \textbf{0.99}   \\
\bottomrule[1.5pt]
\end{tabular}}
\vspace{-10pt}
\label{detection_result_table}
\end{table}

\vspace{-3pt}

\section{Experiment}
\subsection{Experimental Setup}
% 介绍测试数据集，可以说一下正负样本比例
% 为了验证模型的有效性和泛化性，我们选择了一些富有挑战性公开benchmark数据集作为测试集，包括Photoshop篡改数据集（例如CASIAv1，Columbia，Coverage），DeepFake篡改数据集（faceapp，ffhq），另外，为了验证模型对于AIGC-editing的检测定位效果，我们利用AIGC-编辑算法和coco数据，构造了一个aigc检测数据集。测试集的数据来源和正负样本情况如表所示，关于更多细节详见补充材料

% 我们使用3.2节提及的数据集构造方法，分别构造了MMTD-Set的训练集和测试集。我们使用CASIA，Fantasitic Reality(FR)，ffhq，faceapp，coco以及一部分自造aigc数据作为原始训练数据；

\textbf{Dataset:} 
\label{Experimental Setup dataset}
We employ the dataset construction method outlined in Section~\ref{Construction of the proposed MMTD-Set} to build the training and test sets of the MMTD-Set. 
For the training set, we utilize PhotoShop tampering (e.g., CASIAv2~\citep{dong2013casia}, Fantastic Reality~\citep{kniaz2019point}), DeepFake tampering (e.g., FFHQ, FaceApp~\citep{dang2020detectiondffd}), and some self-constructed AIGC-Editing tampered data as the source dataset. For the testing set, we select several challenging public benchmark datasets including PhotoShop tampering (CASIA1+~\citep{dong2013casia}, Columbia~\citep{ng2009columbia}, IMD2020~\citep{novozamsky2020imd2020}, Coverage~\citep{wen2016coverage}, DSO~\citep{de2013exposingdso}, Korus~\citep{korus2016multi}), DeepFake tampering (e.g., FFHQ, FaceApp~\citep{dang2020detectiondffd}, Seq-DeepFake~\citep{shao2022detecting}), and some self-generated AIGC-Editing data.

\textbf{State-of-the-Art Methods:} To ensure a fair comparison, we select competitive methods that provide either open-source code or pre-trained models. To evaluate the \textbf{IFDL performance} of FakeShield, we compare it against SPAN~\citep{hu2020span}, MantraNet~\citep{wu2019mantra}, OSN~\citep{wu2022robust}, HiFi-Net~\citep{guo2023hierarchical}, PSCC-Net~\citep{liu2022pscc}, CAT-Net~\citep{kwon2021cat}, and MVSS-Net~\citep{dong2022mvss}, all of which are retrained on the MMTD-Set for consistency with the same training setup. For \textbf{DeepFake detection}, CADDM~\citep{dong2023implicit}, HiFi-DeepFake~\citep{guo2023hierarchical}, RECCE~\citep{cao2022end} and Exposing~\citep{ba2024exposing} are chosen as comparison methods. Additionally, to assess the \textbf{explanation ability} of FakeShield, we compare it with open-source M-LLMs such as LLaVA-v1.6-34B~\citep{liu2024visual}, InternVL2-26B~\citep{chen2024internvl}, and Qwen2-VL-7B~\citep{wang2024qwen2}, as well as the closed-source model GPT-4o~\citep{openai2023gpt}.
% 介绍对比方法是如何实现的，是用的预训练权重，还是自己在数据集上训的，是用什么数据集训的，怎么测得
% 为了公平的对比，我们选取了一些代码开源或预训练模型可获得的方法。为了评估FakeShield的IFDL性能，我们选择了SPAN，MantraNet，OSN，HiFiNet，PSCC-Net，CAT-Net，MVSS-Net作为对比方法，其中SPAN，MantraNet，OSN，HiFiNet，MVSS-Net使用了作者公布的预训练权重，PSCC-Net，CAT-Net在MMTD-Set数据集上进行了重新训练。为了评估DeepFake检测性能，我们选择CADDM和HiFi-DeepFake。另外，为了评价FakeShield的解释性能，我们选择开源大模型LLaVA-v1.6-34b，InternVL2-26B， Qwen2-VL-7B-Instruct和闭源模型api GPT-4o

% 说一下三个维度的评价指标，检测，解释，定位。各自介绍一下怎么算的，解释那个指标重点讲，写公式
% 对于detection，我们report图像级的acc和F1分数；对于location，我们report IoU和f1分数；对于可解释性，我们使用Cosine Semantic Similarity来评价预测文本与gt文本之间的相似度，通过计算高维语义向量的余弦相似度来实现

\textbf{Evaluation Metrics:} For detection, we report image-level accuracy (ACC) and F1 scores. For localization, we provide Intersection over Union (IoU) and F1 scores. To evaluate interpretability, we use Cosine Semantic Similarity (CSS) to assess the similarity between the predicted text and ground truth text by calculating the cosine similarity between their high-dimensional semantic vectors. For both detection and localization, a default threshold of $0.5$ is applied unless otherwise specified.

\textbf{Implementation Details:} On the MMTD-Set, we initially fine-tune the M-LLM using LoRA (rank=$128$, alpha=$256$), such as LLaVA-v1.5-13B~\citep{liu2024visual}, while simultaneously training the Domain Tag Generator with full parameters. The model is trained for $10$ epochs on $4$ NVIDIA A100 40G GPUs, with a learning rate of $2$$\times$$10^{-4}$. Afterward, we fine-tune the Tamper Comprehension Module and Segment Anything Model~\citep{kirillov2023segment} with LoRA (rank=$8$, alpha=$16$), training for $24$ epochs on the same hardware configuration, with a learning rate of $3$$\times$$10^{-4}$. 
% 为了控制篇幅，这一段尽可能缩减，放在补充材料里面。前三个点要更重要
% 在MMTD-SET上，我们首先用lora（rank=32，alpha=48）微调M-LLM，例如LLaVA1.5，同时全参数训练Domain Tag Generator，我们使用了4台NVIDIA A100 40G训练了4个epoch，学习率设置为2e-4，bs为48
% 随后，我们又用lora微调TCM和SAM，（rank=32，alpha=48），使用4台NVIDIA A100 40G训练了4个epoch，学习率设置为2e-4，bs为48

\subsection{Comparison with Image Forgery Detection Method}
% 为了验证我们的方法在图片篡改检测任务上的优越性与泛化性，我们在Photoshop-tampered datasets， DeepFake-tampered datasets and aigc-editing dataset 上测试了模型性能。如表1所示，我们的FakeShield在各种篡改和各种测试集上几乎都取得了最优的性能。例如，as reported in table 1, 在imd数据集上，我们的方法超过了第二好的方法mvssnet，0.08的acc和0.05的f1. 
% 值得注意的是，由于我们引入了domain tag引导策略，我们的方法不仅在传统的IFDL benchmark上，可以取得优异性能，同时，还可以泛化到DeepFake与aigc数据集上，分别取得0.93和0.98的acc指标。
% 然而，其他工作由于没有良好的处理数据域冲突的机制，即使与我们使用相同的多数据域训练集训练，检测精度仍然不足。

To verify the superiority and generalization of our method on the image forgery detection task, we test the detection accuracy on MMTD-Set (Photoshop, DeepFake, AIGC-Editing). As shown in Table~\ref{detection_result_table}, our FakeShield almost achieves optimal performance across various tampering and testing data domains. For example, our method outperforms the second-best method, MVSS-Net, with an ACC of 0.08 and an F1 of 0.05 on the IMD2020 dataset. Notably, since we introduce the domain-tag guidance strategy, our method not only achieves excellent performance on the traditional IFDL benchmark but also generalizes to DeepFake and AIGC tampering, achieving 0.93 and 0.98 detection accuracy, respectively. However, other works lack an effective mechanism to handle data domain conflicts. Even when trained using the same multi-data domain training set as ours, their detection accuracy remains insufficient.

% \begin{wraptable}{r}{6cm}
% \vspace{-0.3cm}
% % 

%   \newcommand{\tabincell}[2]{\begin{tabular}{@{}#1@{}}#2\end{tabular}}
%   \centering
%   \captionsetup{font={small}}
%     \caption{Performance comparison with competitive DeepFake detection methods.}
%     \vspace{-5pt}
%     \renewcommand{\arraystretch}{1.1}
%   \resizebox{1.\linewidth}{!}{
%     \begin{tabular}{c|ccc}
%     \toprule
%      Method & CADDM &  HiFi-DeepFake &  Ours \\
%     \midrule
%      ACC & \underline{0.5227} & 0.5177 & \textbf{0.9835} \\
%      F1  & 0.5982 & \underline{0.6403} & \textbf{0.9915} \\
%     \bottomrule
%     \end{tabular}}
%     \label{df_detect_result}
%     \vspace{-5pt}
%   \end{wraptable}

\begin{wraptable}{r}{6cm}
\centering
\vspace{-10pt}
\caption{Performance comparison on DFFD and Seq-DeepFake datasets.}
\vspace{-10pt}
\renewcommand{\arraystretch}{1.2}
\resizebox{\linewidth}{!}{
\begin{tabular}{c|cc|cc}
\toprule[1.5pt]
\multirow{2}{*}{Method} & \multicolumn{2}{c|}{DFFD} & \multicolumn{2}{c}{Seq-DeepFake}     \\ \cline{2-5} 
     & ACC  & F1   & ACC  & F1   \\ \hline
CADDM & 0.52 & 0.60 & 0.53 & 0.59 \\ 
HiFi-DeepFake & 0.52 & 0.64 & 0.52 & 0.57 \\ 
Exposing  & 0.82 & 0.84 & 0.71 & 0.78 \\
RECCE  & \underline{0.92} & \underline{0.92} & \underline{0.75} & \underline{0.79} \\ 
\textbf{FakeShield} & \textbf{0.98}     & \textbf{0.99}     & \textbf{0.84}     & \textbf{0.91}     \\ \bottomrule[1.5pt]
\end{tabular}}
\label{tab:comparison_dffd_seqdeepfake}
\end{wraptable}

% 另外，我们还与一些专门进行DeepFake检测的方法在测试集上进行了对比，包括在FaceForensics++ 数据集上预训练的CADDM和HiFi-DeepFake，结果如表1所示，我们的方法大幅领先第二好的方法xxx。值得注意的是，由于我们引入的Domain tag机制，不仅弥合了不同数据域的冲突，不同数据域之间也有一定的互补和正向促进的作用。尽管这些对比方法是专为DeepFake检测设计的，但仍然效果不敌FakeShield   
Furthermore, we compare our method with some recent DeepFake detection approaches on the DFFD (FaceApp, FFHQ) and Seq-DeepFake, where Exposing and RECCE were retrained on our datasets. As reported in Table~\ref{tab:comparison_dffd_seqdeepfake}, our approach significantly outperforms all other methods. Notably, the domain tag mechanism not only alleviates conflicts across diverse data domains but also promotes complementarity and mutual enhancement between them. Although specifically designed for DeepFake detection, these methods still underperform compared to our FakeShield.
\vspace{-5pt}

\begin{figure}[t!]
	\centering
	\includegraphics[width=1\linewidth]{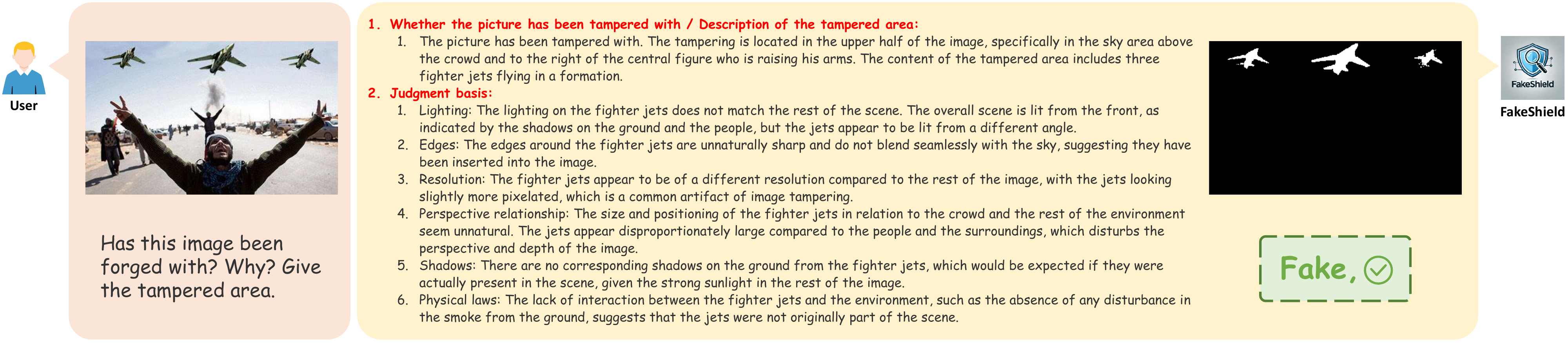}
	\vspace{-18pt}
	\caption{Detection, localization and explanation results of our FakeShield.}
    \vspace{-18pt}
    \label{mainexamplepdf}
\end{figure}

\subsection{Comparison with M-LLMs}
\label{Comparison with M-LLMs}
% 为了评价解释文本生成的质量，我们在MMTD test dataset上，以GPT-4o构造的篡改描述为gt文本，使用CSS评价指标，对主流大语言模型和FakeShield进行了测试。结果如表2所示，我们的方法在几乎所有测试集中，都取得了最好的结果，例如，在DSO数据集中，我们的方法取得了0.8873的css指标，远超第二InternVL0.2389。模型输出部分结果如图所示，值得注意的是，部分预训练M-LLM还是具有优先的篡改检测能力的，例如对于一些由篡改导致的明显违反物理规律的例子，大语言模型可以利用预训练阶段接受的世界知识来正确判断，但其无法进一步的准确的从光照，透视关系等角度来分析篡改图片，精度仍然较低。

To assess the quality of explanation text generation, we employ the tampered descriptions generated by GPT-4o as ground truth on the MMTD-Set(Photoshop, DeepFake, AIGC-Editing), using cosine semantic similarity (CSS) to compare the performance of pre-trained M-LLMs against FakeShield. The results in Table~\ref{explanation_table} show that our approach consistently achieves the best performance across nearly all test sets. For instance, on the DSO, our method attains a CSS score of 0.8873, significantly surpassing the second-best result from InternVL2-26B. Figure~\ref{mainexamplepdf} presents an example of our explanation. The model identifies the three airplanes in the sky as fake and explains its judgment based on lighting and edges, providing an accurate manipulation region mask.
% FakeShield的输出样例如图4所示，模型清楚的指出了天空中的三架飞机是伪造的，并从光影，边缘等角度解释判断依据，给出准确的篡改区域mask

Notably, some pre-trained M-LLMs exhibit limited proficiency in detecting tampered content. For instance, when tampering causes clear physical law violations, these M-LLMs leverage their pre-training knowledge to make reasonably correct judgments. However, they struggle to perform more precise analyses, like detecting lighting or perspective inconsistencies, reducing overall accuracy.

% 预训练M-LLM与FakeShield在MMTD-Set测试的篡改解释能力对比结果。FakeShield在所有测试集中，都处于领先地位，有较强的解释分析能力

\begin{table}[h!]
\centering
\vspace{-5pt}
\caption{Comparative results(CSS$\uparrow$) of the pre-trained M-LLMs and FakeShield in tampering explanation capabilities on the MMTD-Set.}
\vspace{-8pt}
\renewcommand{\arraystretch}{1.3}
\resizebox{1.\linewidth}{!}{
\begin{tabular}{c|ccccccc|c|c}
\toprule
\multirow{2}{*}{Method} & \multicolumn{7}{c|}{PhotoShop} &  \multicolumn{1}{c|}{\multirow{2}{*}{DeepFake}} & \multicolumn{1}{c}{\multirow{2}{*}{AIGC-Editing}} \\
& CASIA1+ & IMD2020 & Columbia & Coverage & NIST & DSO & \multicolumn{1}{c|}{Korus} & \multicolumn{1}{c|}{} & ~ \\
\midrule
% \cmidrule(r){1-1} \cmidrule(r){2-2} \cmidrule(r){3-3} \cmidrule(r){4-4} \cmidrule(r){5-5} \cmidrule(r){6-6} \cmidrule(r){7-7}  \cmidrule(r){8-8} \cmidrule(r){9-9} \cmidrule(r){10-10}

GPT-4o & 0.5183 & 0.5326 & 0.5623 & 0.5518 & 0.5732 & 0.5804 & 0.5549 & 0.5643 & 0.6289 \\
LLaVA-v1.6-34B & 0.6457 & 0.5193 & 0.5578 & 0.5655 & 0.5757 & 0.5034 & 0.5387 & 0.6273 & 0.6352 \\
InternVL2-26B & \underline{0.6760} & \underline{0.5750} & \underline{0.6155} & \underline{0.6193} & \underline{0.6458} & \underline{0.6484} & \underline{0.6297} & \underline{0.6570} & \underline{0.6751} \\
Qwen2-VL-7B & 0.6133 & 0.5351 & 0.5603 & 0.5702 & 0.5559 & 0.4887 & 0.5260 & 0.6060 & 0.6209 \\
FakeShield & \textbf{0.8758} & \textbf{0.7537} & \textbf{0.8791} & \textbf{0.8684} & \textbf{0.8087} & \textbf{0.8873} & \textbf{0.7941} & \textbf{0.8446} & \textbf{0.8860} \\
\bottomrule
\end{tabular}}
\label{explanation_table}
\vspace{-10pt}
\end{table}

\subsection{Comparison with Image Forgery Location Method}
% 为了评价模型的篡改区域定位能力，我们在Photoshop与AIGC-Editing篡改数据集上，与MVSS-Net等IFDL方法进行对比。如表3所示，我们几乎在所有测试集上，都取得了最好的成绩，例如，在IMD2020数据集上，我们的方法在IoU和f1指标上，分别以0.12和0.1的优势，大幅领先第二好的方法OSN。
% 另外，部分方法的主观对比如图4所示，可以看到，我们的方法可以准确的捕捉篡改区域的边缘部分，准确分割出完整的篡改区域。而且部分方法如PSCC-Net对篡改区域的注意力较为分散，且分割较为模糊，预测篡改范围较大。值得注意的是，由于我们的分割模块基于预训练的视觉分割模型SAM，继承其强大的语义分割能力，能够精准的分割具有明确语义信息的目标（如图4的第一第二列），同时，由于得益于MMTD-Set中丰富的篡改类型，对于无语义信息的篡改区域，也能精准分割（如图4的第三列）

To assess the model's capability to locate tampered regions, we conduct comparisons with some competitive IFDL methods on the MMTD-Set. As present in Table~\ref{location_result_table}, our method consistently surpasses other methods across almost all test datasets. For instance, on the IMD2020, our method outperforms the suboptimal method OSN with notable advantages of 0.12 in IoU and 0.1 in F1 score. On the CASIA1+, we also lead OSN with an IoU of 0.07 and an F1 of 0.09.

\begin{figure}[t!]
	\centering
	\includegraphics[width=0.95\linewidth]{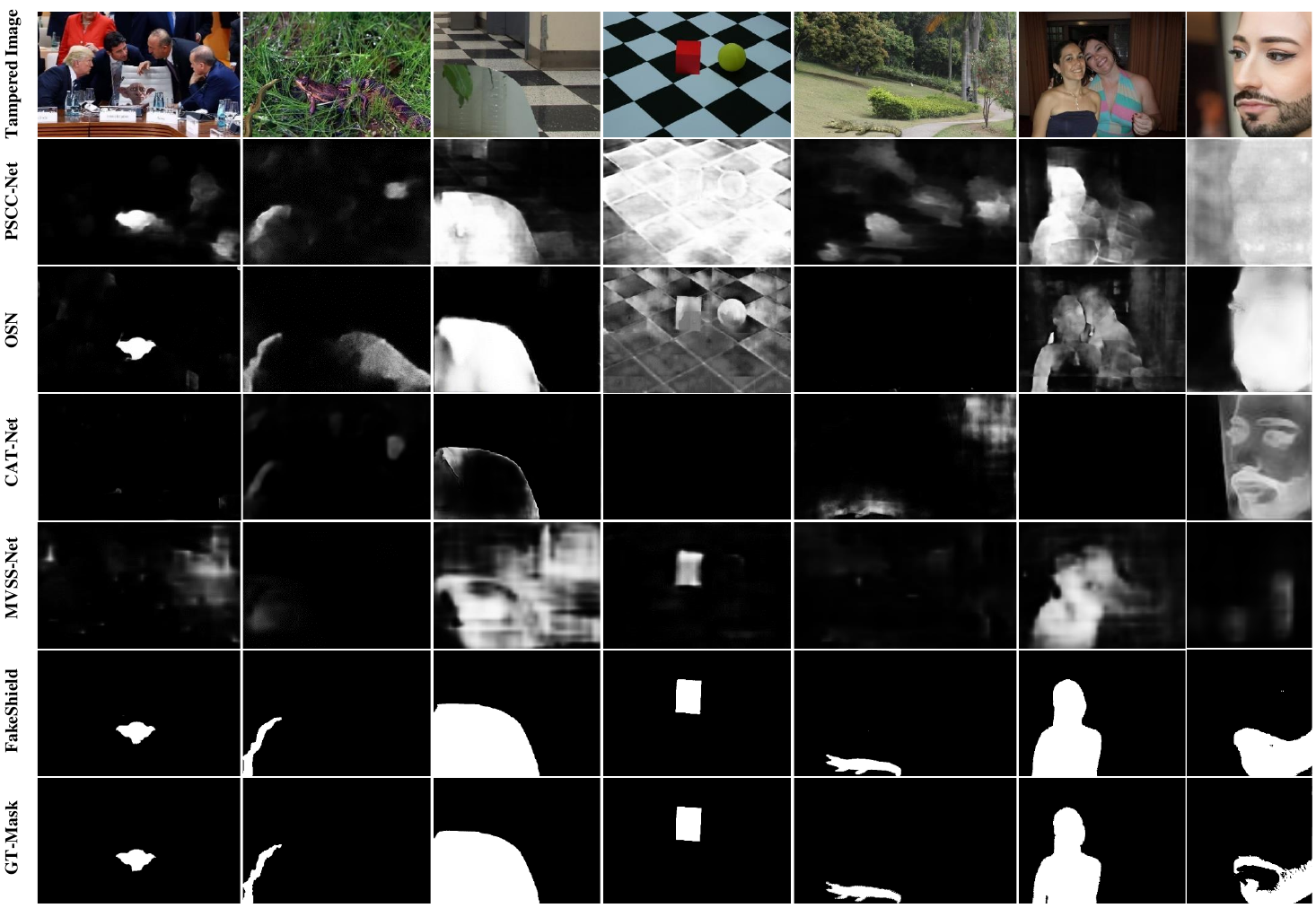}
	\vspace{-5pt}
	\caption{Comparisons between our FakeShield and other competitive methods. The samples, from left to right, are drawn from IMD2020, CASIA1+, Columbia, NIST16, Korus, DSO, and DeepFake.}
    \vspace{-5pt}
    \label{comparation_table}
\end{figure}

Additionally, subjective comparisons of several methods are shown in Figure~\ref{comparation_table}, where our approach precisely captures the tampered areas, producing clean and complete masks. In contrast, methods like PSCC-Net exhibit dispersed attention over the image, resulting in blurred segmentations and an overly broad predicted tampering area. Notably, as our segmentation module MFLM is based on the pre-trained visual model SAM, it inherits SAM's powerful semantic segmentation capabilities and can accurately segment targets with clear semantic information (Columns 1 and 2 of Figure~\ref{comparation_table}). Additionally, due to the diverse tampering types in MMTD-Set, our method can also accurately segmenting tampered areas that lack distinct semantic information (Columns 3 of Figure~\ref{comparation_table}).

% 可比的IFDL方法与FakeShield在MMTD-Set(Photoshop, AIGC-Editing)测试的篡改定位能力对比结果。FakeShield在大多数测试集中，都能取得最高的IoU和f1分数

\begin{table}[t!]
\centering
\caption{Comparative results of tamper localization capabilities between competing IFDL methods and FakeShield, tested on MMTD-Set(Photoshop, DeepFake, AIGC-Editing).}
\vspace{-5pt}
\renewcommand{\arraystretch}{1.2}
\resizebox{1.\linewidth}{!}{
\begin{tabular}{c|cccccccccccccccc}
\toprule
\multirow{2}{*}{\makecell{\centering Method}} & \multicolumn{2}{c}{CASIA1+} & \multicolumn{2}{c}{IMD2020} & \multicolumn{2}{c}{Columbia} & \multicolumn{2}{c}{NIST} & \multicolumn{2}{c}{DSO} & \multicolumn{2}{c}{Korus} & \multicolumn{2}{c}{DeepFake} & \multicolumn{2}{c}{AIGC-Editing} \\
\cmidrule(r){2-3} \cmidrule(r){4-5} \cmidrule(r){6-7} \cmidrule(r){8-9} \cmidrule(r){10-11} \cmidrule(r){12-13} \cmidrule(r){14-15} \cmidrule(r){16-17}
& IoU & F1 & IoU & F1 & IoU & F1 & IoU & F1 & IoU & F1 & IoU & F1 & IoU & F1 & IoU & F1 \\
\midrule
SPAN & 0.11 & 0.14 & 0.09 & 0.14 & 0.14 & 0.20 & 0.16 & 0.21 & 0.14 & 0.24 & 0.06 & 0.10 & 0.04 & 0.06 & 0.09 & 0.12 \\
ManTraNet & 0.09 & 0.13 & 0.10 & 0.16 & 0.04 & 0.07 & 0.14 & 0.20 & 0.08 & 0.13 & 0.02 & 0.05 & 0.03 & 0.05 & 0.07 & 0.12 \\
OSN & \underline{0.47} & \underline{0.51} & \underline{0.38} & \underline{0.47} & 0.58 & 0.69 & \underline{0.25} & \underline{0.33} & \underline{0.32} & \underline{0.45} & 0.14 & 0.19 & 0.11 & 0.13 & 0.07 & 0.09 \\
HiFi-Net & 0.13 & 0.18 & 0.09 & 0.14 & 0.06 & 0.11 & 0.09 & 0.13 & 0.18 & 0.29 & 0.01 & 0.02 & 0.07 & 0.11 & \underline{0.13} & \underline{0.22}  \\
PSCC-Net & 0.36 & 0.46 & 0.22 & 0.32 & \underline{0.64} & \underline{0.74} & 0.18 & 0.26 & 0.22 & 0.33 & \underline{0.15} & \textbf{0.22} & \underline{0.12} & \underline{0.18} & 0.10 & 0.15 \\
CAT-Net & 0.44  & \underline{0.51}  & 0.14  & 0.19  & 0.08  & 0.13  & 0.14  & 0.19  & 0.06  & 0.10  & 0.04  & 0.06 & 0.10 & 0.15 & 0.03  & 0.05  \\
MVSS-Net & 0.40  & 0.48  & 0.23  & 0.31  & 0.48  & 0.61  & 0.24  & 0.29  & 0.23  & 0.34  & 0.12  & 0.17 & 0.10 & 0.09  & \textbf{0.18}  & \textbf{0.24} \\
FakeShield & \textbf{0.54} & \textbf{0.60} & \textbf{0.50} & \textbf{0.57} & \textbf{0.67} & \textbf{0.75} & \textbf{0.32} & \textbf{0.37} & \textbf{0.48} & \textbf{0.52} & \textbf{0.17} & \underline{0.20} & \textbf{0.14} & \textbf{0.22} & \textbf{0.18} & \textbf{0.24} \\
\bottomrule
\end{tabular}}
\label{location_result_table}
\vspace{-5pt}
\end{table}

\subsection{Robustness Study}
% 由于网络和社交媒体的普及，当前人们接收大量由社交媒体传输导致的退化的图片，例如jpeg压缩和高斯噪声。我们报告了模型在面对上述退化时的检测，解释，定位性能。结果如表1所示。可以看到，由于大语言模型更多的关注高维的语义信息，尽管，我们并没有在训练过程中特意加入退化数据进行训练，FakeShield对于底层视觉的干扰和噪声并不敏感，在社交媒体常见的jpeg压缩和高斯噪声下性能影响很小，展现了方法的稳定性，鲁棒性和实用性

With the widespread use of the Internet and social media, individuals are increasingly receiving images degraded by transmission artifacts such as JPEG compression and Gaussian noise. Our model’s performance on MMTD-Set(CASIA1+) under these degradations is reported in Table~\ref{degrade}, which includes four common degradation types: JPEG compression qualities of $70$ and $80$, and Gaussian noise variances of $5$ and $10$. As M-LLMs primarily emphasize high-level semantic information, although we do not specifically add degraded data during training, FakeShield demonstrates robustness to low-level visual distortions and noise. JPEG compression and Gaussian noise, commonly associated with social media, have minimal effect on its performance, highlighting the stability, robustness, and practical advantages of our approach.

% \vspace{-10pt}

\begin{center}
\makeatletter\def\@captype{table}\makeatother
\begin{minipage}{0.45\textwidth}
% \vspace{-0.3cm}
\centering
\caption{Explanation and location performance under different degradations.}
\label{degrade}
\vspace{-10pt}
\resizebox{1.\linewidth}{!}{
\begin{tabular}{c|c|cc}
\toprule
\multirow{2}{*}{Method} & \multicolumn{1}{c|}{Explanation} & \multicolumn{2}{c}{Location} \\
\cmidrule(r){2-2} \cmidrule(r){3-4} 
~ & CSS & IoU & F1 \\
\midrule
JPEG 70 & 0.8355 & 0.5022 & 0.5645 \\
JPEG 80 & 0.8511 & 0.5026 & 0.5647 \\
Gaussian 5 & 0.8283 & 0.4861 & 0.5494 \\
Gaussian 10 & 0.8293 & 0.4693 & 0.5297 \\
Original & 0.8758 & 0.5432 & 0.6032 \\
\bottomrule
\end{tabular}}
\end{minipage}
\hspace{0.3cm}
\makeatletter\def\@captype{figure}\makeatother
\begin{minipage}{0.45\textwidth}
\centering
  \includegraphics[width=0.95\linewidth]{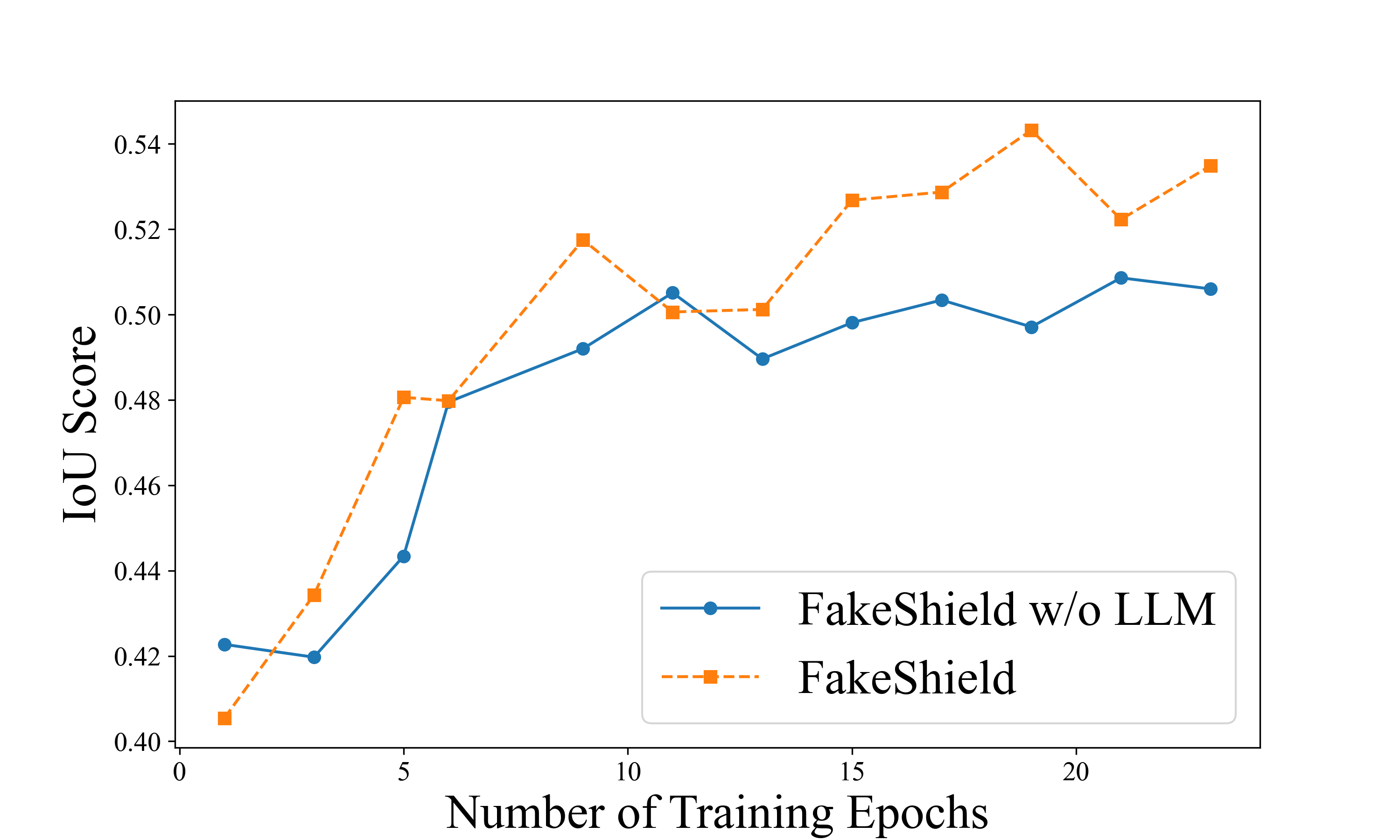}
  \vspace{-8pt}
  \caption{Ablation study on the DTE-FDM.}
  \label{ablation2}
\end{minipage}
\end{center}

% \vspace{-10pt}
\subsection{Ablation Study}
% 为了验证我们提出的domain tag可以有效的弥合数据域差异，提升模型在不同数据域上的泛化性，我们对其进行了消融实验。我们去除Domain Tag Generator模块，在MMTD-Set 训练数据集上对FakeShield进行相同配置的训练。结果如表1所示，去除DTG模块后，模型的检测能力在各个数据域的测试集上都又一定程度下降，其中，DeepFake检测能力，在acc和f1指标上都下降了0.1。由此可知，失去了DTG的辅助，模型难以有效区分不同数据域，数据冲突现象较为严重，泛化性与实用性大幅下降

% 除此之外，为了验证DTE-FDM的必要性，我们进行了进一步的实验。将Tins，Timg直接输入Tamper Comprehension Module，将其训练目标设计为直接输出Odet和<seg>，并提供给sam输出mask。我们对Tamper Comprehension Module使用同样的训练配置，训练的是个epoch，在CASIA1+数据集上进行测试location准确性，结果如表1所示。可以看出，在去除LLM模块后，在整个训练过程中，其定位性能几乎一直低于原版，而且模型收敛更快
% 。由于定位辅助和图像理解任务跨度较大，让同一个模型兼具两种不同的功能，会带来较大的性能损失，证明了LLM模块对提升性能的关键作用。

\textbf{Ablation Study on Domain Tag Generator:} To validate that the proposed domain tag effectively mitigates domain discrepancies and enhances the model's generalization across diverse data domains, we conducted an ablation study. Specifically, we removed the domain tag generator (DTG) and trained the FakeShield with identical configurations on the MMTD-Set, the test results are displayed in Table~\ref{ablation_DTG_table}.
Without the DTG, the model's detection performance declined across test sets from each data domain. Notably, the detection ACC and F1 score for IMD2020 decreased by 0.12 and 0.11. This demonstrates that without the support of the DTG module, the model struggles to effectively differentiate between various data domains, leading to more pronounced data conflicts and a significant reduction in both generalization and practical applicability.

\begin{wraptable}{r}{7cm}
\vspace{-0.3cm}
  \newcommand{\tabincell}[2]{\begin{tabular}{@{}#1@{}}#2\end{tabular}}
\centering
\caption{Performance comparison of FakeShield with and without DTG on different datasets.}
\vspace{-10pt}
\renewcommand{\arraystretch}{1.2}
\resizebox{1.\linewidth}{!}{
\begin{tabular}{c|cccccccccccccc}
\toprule
\multirow{2}{*}{Method} & \multicolumn{2}{c}{CASIA1+} & \multicolumn{2}{c}{IMD2020} & \multicolumn{2}{c}{DeepFake} & \multicolumn{2}{c}{AIGC-Editing} \\
\cmidrule(r){2-3} \cmidrule(r){4-5} \cmidrule(r){6-7} \cmidrule(r){8-9}
& ACC & F1 & ACC & F1 & ACC & F1 & ACC & F1 \\
\midrule
Ours w/o DTG & 0.92 & 0.92 & 0.71 & 0.79 & 0.89 & 0.90 & 0.72 & 0.78 \\
Ours & \textbf{0.95} & \textbf{0.95} & \textbf{0.83} & \textbf{0.90} & \textbf{0.98} & \textbf{0.99} & \textbf{0.93} & \textbf{0.93} \\
\bottomrule
\end{tabular}}
\vspace{-10pt}
\label{ablation_DTG_table}
\end{wraptable}

\textbf{Ablation Study on LLM in the DTE-FDM:} Furthermore, to verify the necessity of the LLM in the DTE-FDM, we design a variant of FakeShield, which removes the LLM and directly input \{$\mathbf{T}_{ins}$, $\mathbf{T}_{tag}$, $\mathbf{T}_{img}$\} into the tamper comprehension module, adjusting its training objective to directly produce the description $\mathbf{O}_{det}$ and the mask $\mathbf{M}_{loc}$. Using the same training configurations, we train the variant and our original framework for $25$ epochs, evaluating their localization accuracy on the CASIA1+ dataset.
As shown in Figure~\ref{ablation2}, after removing the LLM, the localization performance consistently lags behind the original framework throughout the entire training process and it converges earlier. It proves that joint training detection and localization via a single MFLM tends to cause notable performance degradation than our decoupled module design, which further highlights the critical role of our LLM module in enhancing the semantic understanding of the proposed framework.

\section{Conclution}
In this work, we present the first application of an M-LLM for the explanation IFDL, marking a significant advancement in the field. Our proposed framework, FakeShield, excels in tampering detection while delivering comprehensive explanations and precise localization, demonstrating strong generalization across a wide range of manipulation types. These features make it a versatile and practical tool for diverse real-world applications. Looking to the future, this work can play a crucial role in several areas, such as aiding in the improvement of laws and regulations related to digital content manipulation, informing the development of guidelines for generative artificial intelligence, and promoting a clearer and more trustworthy online environment. Additionally, FakeShield can assist in evidence collection for legal proceedings and help correct misinformation in public discourse, ultimately contributing to the integrity and reliability of digital media.

% \section{Acknowledge}
% This work was supported by National Natural Science Foundation of China under Grant 62372016. (Corresponding author: Jian Zhang.)

\clearpage
% 第一个大模型工作，
% 良好的效果，全能
% 展望未来，对社区的作用，工作的价值

\bibliography{iclr2025_conference}

\begin{thebibliography}{68}
\providecommand{\natexlab}[1]{#1}
\providecommand{\url}[1]{\texttt{#1}}
\expandafter\ifx\csname urlstyle\endcsname\relax
  \providecommand{\doi}[1]{doi: #1}\else
  \providecommand{\doi}{doi: \begingroup \urlstyle{rm}\Url}\fi

\bibitem[Asnani et~al.(2023)Asnani, Yin, Hassner, and Liu]{asnani2023malp}
Vishal Asnani, Xi~Yin, Tal Hassner, and Xiaoming Liu.
\newblock Malp: Manipulation localization using a proactive scheme.
\newblock In \emph{Proceedings of the IEEE/CVF Conference on Computer Vision and Pattern Recognition (CVPR)}, 2023.

\bibitem[Ba et~al.(2024)Ba, Liu, Liu, Wu, Lin, Lu, and Ren]{ba2024exposing}
Zhongjie Ba, Qingyu Liu, Zhenguang Liu, Shuang Wu, Feng Lin, Li~Lu, and Kui Ren.
\newblock Exposing the deception: Uncovering more forgery clues for deepfake detection.
\newblock In \emph{Proceedings of the AAAI Conference on Artificial Intelligence}, volume~38, pp.\  719--728, 2024.

\bibitem[Cao et~al.(2022)Cao, Ma, Yao, Chen, Ding, and Yang]{cao2022end}
Junyi Cao, Chao Ma, Taiping Yao, Shen Chen, Shouhong Ding, and Xiaokang Yang.
\newblock End-to-end reconstruction-classification learning for face forgery detection.
\newblock In \emph{Proceedings of the IEEE/CVF Conference on Computer Vision and Pattern Recognition}, pp.\  4113--4122, 2022.

\bibitem[Chen et~al.(2023{\natexlab{a}})Chen, Li, Dong, Zhang, He, Wang, Zhao, and Lin]{chen2023sharegpt4v}
Lin Chen, Jisong Li, Xiaoyi Dong, Pan Zhang, Conghui He, Jiaqi Wang, Feng Zhao, and Dahua Lin.
\newblock Sharegpt4v: Improving large multi-modal models with better captions.
\newblock \emph{arXiv preprint arXiv:2311.12793}, 2023{\natexlab{a}}.

\bibitem[Chen et~al.(2021)Chen, Dong, Ji, Cao, and Li]{chen2021image}
Xinru Chen, Chengbo Dong, Jiaqi Ji, Juan Cao, and Xirong Li.
\newblock Image manipulation detection by multi-view multi-scale supervision.
\newblock In \emph{Proceedings of the IEEE/CVF International Conference on Computer Vision (ICCV)}, 2021.

\bibitem[Chen et~al.(2023{\natexlab{b}})Chen, Wang, Xing, Xu, Fang, Wang, Li, Wu, Liu, Xu, et~al.]{chen2023bianque}
Yirong Chen, Zhenyu Wang, Xiaofen Xing, Zhipei Xu, Kai Fang, Junhong Wang, Sihang Li, Jieling Wu, Qi~Liu, Xiangmin Xu, et~al.
\newblock Bianque: Balancing the questioning and suggestion ability of health llms with multi-turn health conversations polished by chatgpt.
\newblock \emph{arXiv preprint arXiv:2310.15896}, 2023{\natexlab{b}}.

\bibitem[Chen et~al.(2024)Chen, Wu, Wang, Su, Chen, Xing, Zhong, Zhang, Zhu, Lu, et~al.]{chen2024internvl}
Zhe Chen, Jiannan Wu, Wenhai Wang, Weijie Su, Guo Chen, Sen Xing, Muyan Zhong, Qinglong Zhang, Xizhou Zhu, Lewei Lu, et~al.
\newblock Internvl: Scaling up vision foundation models and aligning for generic visual-linguistic tasks.
\newblock In \emph{Proceedings of the IEEE/CVF Conference on Computer Vision and Pattern Recognition}, pp.\  24185--24198, 2024.

\bibitem[Dang et~al.(2020)Dang, Liu, Stehouwer, Liu, and Jain]{dang2020detectiondffd}
Hao Dang, Feng Liu, Joel Stehouwer, Xiaoming Liu, and Anil~K Jain.
\newblock On the detection of digital face manipulation.
\newblock In \emph{Proceedings of the IEEE/CVF Conference on Computer Vision and Pattern recognition}, pp.\  5781--5790, 2020.

\bibitem[De~Carvalho et~al.(2013)De~Carvalho, Riess, Angelopoulou, Pedrini, and de~Rezende~Rocha]{de2013exposingdso}
Tiago~Jos{\'e} De~Carvalho, Christian Riess, Elli Angelopoulou, Helio Pedrini, and Anderson de~Rezende~Rocha.
\newblock Exposing digital image forgeries by illumination color classification.
\newblock \emph{IEEE Transactions on Information Forensics and Security}, 8\penalty0 (7):\penalty0 1182--1194, 2013.

\bibitem[Devlin(2018)]{devlin2018bert}
Jacob Devlin.
\newblock Bert: Pre-training of deep bidirectional transformers for language understanding.
\newblock \emph{arXiv preprint arXiv:1810.04805}, 2018.

\bibitem[Dong et~al.(2022)Dong, Chen, Hu, Cao, and Li]{dong2022mvss}
Chengbo Dong, Xinru Chen, Ruohan Hu, Juan Cao, and Xirong Li.
\newblock Mvss-net: Multi-view multi-scale supervised networks for image manipulation detection.
\newblock \emph{IEEE Transactions on Pattern Analysis and Machine Intelligence}, 45\penalty0 (3):\penalty0 3539--3553, 2022.

\bibitem[Dong et~al.(2013)Dong, Wang, and Tan]{dong2013casia}
Jing Dong, Wei Wang, and Tieniu Tan.
\newblock Casia image tampering detection evaluation database.
\newblock In \emph{Proceedings of the IEEE China Summit and International Conference on Signal and Information Processing (ChinaSIP)}, 2013.

\bibitem[Dong et~al.(2023)Dong, Wang, Ji, Liang, Fan, and Ge]{dong2023implicit}
Shichao Dong, Jin Wang, Renhe Ji, Jiajun Liang, Haoqiang Fan, and Zheng Ge.
\newblock Implicit identity leakage: The stumbling block to improving deepfake detection generalization.
\newblock In \emph{Proceedings of the IEEE/CVF Conference on Computer Vision and Pattern Recognition}, pp.\  3994--4004, 2023.

\bibitem[Dubey et~al.(2024)Dubey, Jauhri, Pandey, Kadian, Al-Dahle, Letman, Mathur, Schelten, Yang, Fan, et~al.]{dubey2024llama}
Abhimanyu Dubey, Abhinav Jauhri, Abhinav Pandey, Abhishek Kadian, Ahmad Al-Dahle, Aiesha Letman, Akhil Mathur, Alan Schelten, Amy Yang, Angela Fan, et~al.
\newblock The llama 3 herd of models.
\newblock \emph{arXiv preprint arXiv:2407.21783}, 2024.

\bibitem[{FaceApp Limited}(2017)]{faceapp2017}
{FaceApp Limited}.
\newblock Faceapp.
\newblock \url{https://www.faceapp.com/}, 2017.
\newblock Accessed: 2024-08-16.

\bibitem[Guo et~al.(2023)Guo, Liu, Ren, Grosz, Masi, and Liu]{guo2023hierarchical}
Xiao Guo, Xiaohong Liu, Zhiyuan Ren, Steven Grosz, Iacopo Masi, and Xiaoming Liu.
\newblock Hierarchical fine-grained image forgery detection and localization.
\newblock In \emph{Proceedings of the IEEE/CVF Conference on Computer Vision and Pattern Recognition (CVPR)}, 2023.

\bibitem[Hu et~al.(2022)Hu, belong shen, Wallis, Allen-Zhu, Li, Wang, Wang, and Chen]{hu2022lora}
Edward~J Hu, belong shen, Phillip Wallis, Zeyuan Allen-Zhu, Yuanzhi Li, Shean Wang, Lu~Wang, and Weizhu Chen.
\newblock Lo{RA}: Low-rank adaptation of large language models.
\newblock In \emph{International Conference on Learning Representations (ICLR)}, 2022.

\bibitem[Hu et~al.(2023)Hu, Ying, Qian, Li, and Zhang]{hu2023draw}
Xiaoxiao Hu, Qichao Ying, Zhenxing Qian, Sheng Li, and Xinpeng Zhang.
\newblock Draw: Defending camera-shooted raw against image manipulation.
\newblock In \emph{Proceedings of the IEEE/CVF International Conference on Computer Vision (ICCV)}, 2023.

\bibitem[Hu et~al.(2020)Hu, Zhang, Jiang, Chaudhuri, Yang, and Nevatia]{hu2020span}
Xuefeng Hu, Zhihan Zhang, Zhenye Jiang, Syomantak Chaudhuri, Zhenheng Yang, and Ram Nevatia.
\newblock Span: Spatial pyramid attention network for image manipulation localization.
\newblock In \emph{Proceedings of the European Conference on Computer Vision (ECCV)}, 2020.

\bibitem[Huang et~al.(2024)Huang, Xia, Lin, Mou, and Yang]{huang2024ffaa}
Zhengchao Huang, Bin Xia, Zicheng Lin, Zhun Mou, and Wenming Yang.
\newblock Ffaa: Multimodal large language model based explainable open-world face forgery analysis assistant.
\newblock \emph{arXiv preprint arXiv:2408.10072}, 2024.

\bibitem[Islam et~al.(2020)Islam, Long, Basharat, and Hoogs]{islam2020doa}
Ashraful Islam, Chengjiang Long, Arslan Basharat, and Anthony Hoogs.
\newblock Doa-gan: Dual-order attentive generative adversarial network for image copy-move forgery detection and localization.
\newblock In \emph{Proceedings of the IEEE/CVF Conference on Computer Vision and Pattern Recognition (CVPR)}, 2020.

\bibitem[Kirillov et~al.(2023)Kirillov, Mintun, Ravi, Mao, Rolland, Gustafson, Xiao, Whitehead, Berg, Lo, et~al.]{kirillov2023segment}
Alexander Kirillov, Eric Mintun, Nikhila Ravi, Hanzi Mao, Chloe Rolland, Laura Gustafson, Tete Xiao, Spencer Whitehead, Alexander~C Berg, Wan-Yen Lo, et~al.
\newblock Segment anything.
\newblock \emph{arXiv preprint arXiv:2304.02643}, 2023.

\bibitem[Kniaz et~al.(2019)Kniaz, Knyaz, and Remondino]{kniaz2019point}
Vladimir~V Kniaz, Vladimir Knyaz, and Fabio Remondino.
\newblock The point where reality meets fantasy: Mixed adversarial generators for image splice detection.
\newblock \emph{Advances in neural information processing systems}, 32, 2019.

\bibitem[Korus \& Huang(2016)Korus and Huang]{korus2016multi}
Pawe{\l} Korus and Jiwu Huang.
\newblock Multi-scale analysis strategies in prnu-based tampering localization.
\newblock \emph{IEEE Transactions on Information Forensics and Security}, 12\penalty0 (4):\penalty0 809--824, 2016.

\bibitem[Kwon et~al.(2021)Kwon, Yu, Nam, and Lee]{kwon2021cat}
Myung-Joon Kwon, In-Jae Yu, Seung-Hun Nam, and Heung-Kyu Lee.
\newblock Cat-net: Compression artifact tracing network for detection and localization of image splicing.
\newblock In \emph{Proceedings of the IEEE/CVF Winter Conference on Applications of Computer Vision (WACV)}, 2021.

\bibitem[Lai et~al.(2024)Lai, Tian, Chen, Li, Yuan, Liu, and Jia]{lai2024lisa}
Xin Lai, Zhuotao Tian, Yukang Chen, Yanwei Li, Yuhui Yuan, Shu Liu, and Jiaya Jia.
\newblock Lisa: Reasoning segmentation via large language model.
\newblock In \emph{Proceedings of the IEEE/CVF Conference on Computer Vision and Pattern Recognition}, pp.\  9579--9589, 2024.

\bibitem[Li \& Huang(2019)Li and Huang]{li2019localization}
Haodong Li and Jiwu Huang.
\newblock Localization of deep inpainting using high-pass fully convolutional network.
\newblock In \emph{Proceedings of the IEEE/CVF International Conference on Computer Vision (ICCV)}, 2019.

\bibitem[Li et~al.(2022)Li, Li, Xiong, and Hoi]{li2022blip}
Junnan Li, Dongxu Li, Caiming Xiong, and Steven Hoi.
\newblock Blip: Bootstrapping language-image pre-training for unified vision-language understanding and generation.
\newblock In \emph{International Conference on Machine Learning (ICML)}, 2022.

\bibitem[Li et~al.(2024)Li, Zhang, Xu, Zhang, and Zhang]{li2024protect}
Runyi Li, Xuanyu Zhang, Zhipei Xu, Yongbing Zhang, and Jian Zhang.
\newblock Protect-your-ip: Scalable source-tracing and attribution against personalized generation.
\newblock \emph{arXiv preprint arXiv:2405.16596}, 2024.

\bibitem[Li \& Zhou(2018)Li and Zhou]{li2018fast}
Yuanman Li and Jiantao Zhou.
\newblock Fast and effective image copy-move forgery detection via hierarchical feature point matching.
\newblock \emph{IEEE Transactions on Information Forensics and Security}, 14\penalty0 (5):\penalty0 1307--1322, 2018.

\bibitem[Li et~al.(2018)Li, Liu, Li, Li, Li, and Wu]{li2018learning}
Yue Li, Dong Liu, Houqiang Li, Li~Li, Zhu Li, and Feng Wu.
\newblock Learning a convolutional neural network for image compact-resolution.
\newblock \emph{IEEE Transactions on Image Processing}, 28\penalty0 (3):\penalty0 1092--1107, 2018.

\bibitem[Lin et~al.(2014)Lin, Maire, Belongie, Hays, Perona, Ramanan, Doll{\'a}r, and Zitnick]{lin2014microsoft}
Tsung-Yi Lin, Michael Maire, Serge Belongie, James Hays, Pietro Perona, Deva Ramanan, Piotr Doll{\'a}r, and C~Lawrence Zitnick.
\newblock Microsoft coco: Common objects in context.
\newblock In \emph{Proceedings of the European Conference on Computer Vision (ECCV)}, 2014.

\bibitem[Liu et~al.(2024)Liu, Li, Wu, and Lee]{liu2024visual}
Haotian Liu, Chunyuan Li, Qingyang Wu, and Yong~Jae Lee.
\newblock Visual instruction tuning.
\newblock \emph{Advances in neural information processing systems}, 36, 2024.

\bibitem[Liu et~al.(2022)Liu, Liu, Chen, and Liu]{liu2022pscc}
Xiaohong Liu, Yaojie Liu, Jun Chen, and Xiaoming Liu.
\newblock Pscc-net: Progressive spatio-channel correlation network for image manipulation detection and localization.
\newblock \emph{IEEE Transactions on Circuits and Systems for Video Technology}, 32\penalty0 (11):\penalty0 7505--7517, 2022.

\bibitem[Lugmayr et~al.(2022)Lugmayr, Danelljan, Romero, Yu, Timofte, and Van~Gool]{Lugmayr_2022_CVPR}
Andreas Lugmayr, Martin Danelljan, Andres Romero, Fisher Yu, Radu Timofte, and Luc Van~Gool.
\newblock Repaint: Inpainting using denoising diffusion probabilistic models.
\newblock In \emph{Proceedings of the IEEE/CVF Conference on Computer Vision and Pattern Recognition (CVPR)}, 2022.

\bibitem[Ma et~al.(2023)Ma, Du, Liu, Hammadi, and Zhou]{ma2023iml}
Xiaochen Ma, Bo~Du, Xianggen Liu, Ahmed Y~Al Hammadi, and Jizhe Zhou.
\newblock Iml-vit: Image manipulation localization by vision transformer.
\newblock \emph{arXiv preprint arXiv:2307.14863}, 2023.

\bibitem[Mou et~al.(2023)Mou, Wang, Song, Shan, and Zhang]{mou2023dragondiffusion}
Chong Mou, Xintao Wang, Jiechong Song, Ying Shan, and Jian Zhang.
\newblock Dragondiffusion: Enabling drag-style manipulation on diffusion models.
\newblock \emph{arXiv preprint arXiv:2307.02421}, 2023.

\bibitem[Ng et~al.(2009)Ng, Hsu, and Chang]{ng2009columbia}
Tian-Tsong Ng, Jessie Hsu, and Shih-Fu Chang.
\newblock Columbia image splicing detection evaluation dataset.
\newblock \emph{DVMM lab. Columbia Univ CalPhotos Digit Libr}, 2009.

\bibitem[Nirkin et~al.(2021)Nirkin, Wolf, Keller, and Hassner]{nirkin2021DeepFake}
Yuval Nirkin, Lior Wolf, Yosi Keller, and Tal Hassner.
\newblock Deepfake detection based on discrepancies between faces and their context.
\newblock \emph{IEEE Transactions on Pattern Analysis and Machine Intelligence}, 44\penalty0 (10):\penalty0 6111--6121, 2021.

\bibitem[Novozamsky et~al.(2020)Novozamsky, Mahdian, and Saic]{novozamsky2020imd2020}
Adam Novozamsky, Babak Mahdian, and Stanislav Saic.
\newblock Imd2020: A large-scale annotated dataset tailored for detecting manipulated images.
\newblock In \emph{Proceedings of the IEEE/CVF Winter Conference on Applications of Computer Vision Workshops}, pp.\  71--80, 2020.

\bibitem[{NVIDIA Corporation}()]{ffhqdataset}
{NVIDIA Corporation}.
\newblock Flickr-faces-hq dataset (ffhq).
\newblock \url{https://github.com/NVlabs/ffhq-dataset}.
\newblock Accessed: 2024-09-30.

\bibitem[OpenAI(2023)]{openai2023gpt}
R~OpenAI.
\newblock Gpt-4 technical report. arxiv 2303.08774.
\newblock \emph{View in Article}, 2\penalty0 (5), 2023.

\bibitem[Podell et~al.(2023)Podell, English, Lacey, Blattmann, Dockhorn, M{\"u}ller, Penna, and Rombach]{podell2023sdxl}
Dustin Podell, Zion English, Kyle Lacey, Andreas Blattmann, Tim Dockhorn, Jonas M{\"u}ller, Joe Penna, and Robin Rombach.
\newblock Sdxl: improving latent diffusion models for high-resolution image synthesis.
\newblock \emph{arXiv preprint arXiv:2307.01952}, 2023.

\bibitem[Rasheed et~al.(2024)Rasheed, Maaz, Shaji, Shaker, Khan, Cholakkal, Anwer, Xing, Yang, and Khan]{rasheed2024glamm}
Hanoona Rasheed, Muhammad Maaz, Sahal Shaji, Abdelrahman Shaker, Salman Khan, Hisham Cholakkal, Rao~M Anwer, Eric Xing, Ming-Hsuan Yang, and Fahad~S Khan.
\newblock Glamm: Pixel grounding large multimodal model.
\newblock In \emph{Proceedings of the IEEE/CVF Conference on Computer Vision and Pattern Recognition}, pp.\  13009--13018, 2024.

\bibitem[Rombach et~al.(2022)Rombach, Blattmann, Lorenz, Esser, and Ommer]{rombach2022high}
Robin Rombach, Andreas Blattmann, Dominik Lorenz, Patrick Esser, and Bj{\"o}rn Ommer.
\newblock High-resolution image synthesis with latent diffusion models.
\newblock In \emph{Proceedings of the IEEE/CVF Conference on Computer Vision and Pattern Recognition (CVPR)}, 2022.

\bibitem[Salloum et~al.(2018)Salloum, Ren, and Kuo]{salloum2018image}
Ronald Salloum, Yuzhuo Ren, and C-C~Jay Kuo.
\newblock Image splicing localization using a multi-task fully convolutional network (mfcn).
\newblock \emph{Journal of Visual Communication and Image Representation}, 51:\penalty0 201--209, 2018.

\bibitem[Sanh et~al.(2022)Sanh, Webson, Raffel, Bach, Sutawika, Alyafeai, Chaffin, Stiegler, Raja, Dey, et~al.]{sanh2022multitask}
Victor Sanh, Albert Webson, Colin Raffel, Stephen Bach, Lintang Sutawika, Zaid Alyafeai, Antoine Chaffin, Arnaud Stiegler, Arun Raja, Manan Dey, et~al.
\newblock Multitask prompted training enables zero-shot task generalization.
\newblock In \emph{International Conference on Learning Representations}, 2022.

\bibitem[Shao et~al.(2022)Shao, Wu, and Liu]{shao2022detecting}
Rui Shao, Tianxing Wu, and Ziwei Liu.
\newblock Detecting and recovering sequential deepfake manipulation.
\newblock In \emph{European Conference on Computer Vision}, pp.\  712--728. Springer, 2022.

\bibitem[Sudre et~al.(2017)Sudre, Li, Vercauteren, Ourselin, and Jorge~Cardoso]{sudre2017generalised}
Carole~H Sudre, Wenqi Li, Tom Vercauteren, Sebastien Ourselin, and M~Jorge~Cardoso.
\newblock Generalised dice overlap as a deep learning loss function for highly unbalanced segmentations.
\newblock In \emph{Deep Learning in Medical Image Analysis and Multimodal Learning for Clinical Decision Support: Third International Workshop, DLMIA 2017, and 7th International Workshop, ML-CDS 2017, Held in Conjunction with MICCAI 2017, Qu{\'e}bec City, QC, Canada, September 14, Proceedings 3}, pp.\  240--248. Springer, 2017.

\bibitem[Suvorov et~al.(2022)Suvorov, Logacheva, Mashikhin, Remizova, Ashukha, Silvestrov, Kong, Goka, Park, and Lempitsky]{suvorov2022resolution}
Roman Suvorov, Elizaveta Logacheva, Anton Mashikhin, Anastasia Remizova, Arsenii Ashukha, Aleksei Silvestrov, Naejin Kong, Harshith Goka, Kiwoong Park, and Victor Lempitsky.
\newblock Resolution-robust large mask inpainting with fourier convolutions.
\newblock In \emph{Proceedings of the IEEE/CVF Winter Conference on Applications of Computer Vision (WACV)}, 2022.

\bibitem[Wang et~al.(2024)Wang, Bai, Tan, Wang, Fan, Bai, Chen, Liu, Wang, Ge, et~al.]{wang2024qwen2}
Peng Wang, Shuai Bai, Sinan Tan, Shijie Wang, Zhihao Fan, Jinze Bai, Keqin Chen, Xuejing Liu, Jialin Wang, Wenbin Ge, et~al.
\newblock Qwen2-vl: Enhancing vision-language model's perception of the world at any resolution.
\newblock \emph{arXiv preprint arXiv:2409.12191}, 2024.

\bibitem[Wang et~al.(2023)Wang, Lv, Yu, Hong, Qi, Wang, Ji, Yang, Zhao, Song, et~al.]{wang2023cogvlm}
Weihan Wang, Qingsong Lv, Wenmeng Yu, Wenyi Hong, Ji~Qi, Yan Wang, Junhui Ji, Zhuoyi Yang, Lei Zhao, Xixuan Song, et~al.
\newblock Cogvlm: Visual expert for pretrained language models.
\newblock \emph{arXiv preprint arXiv:2311.03079}, 2023.

\bibitem[Wei et~al.(2022)Wei, Wang, Schuurmans, Bosma, Xia, Chi, Le, Zhou, et~al.]{wei2022chain}
Jason Wei, Xuezhi Wang, Dale Schuurmans, Maarten Bosma, Fei Xia, Ed~Chi, Quoc~V Le, Denny Zhou, et~al.
\newblock Chain-of-thought prompting elicits reasoning in large language models.
\newblock \emph{Advances in neural information processing systems}, 35:\penalty0 24824--24837, 2022.

\bibitem[Wen et~al.(2016)Wen, Zhu, Subramanian, Ng, Shen, and Winkler]{wen2016coverage}
Bihan Wen, Ye~Zhu, Ramanathan Subramanian, Tian-Tsong Ng, Xuanjing Shen, and Stefan Winkler.
\newblock Coverage—a novel database for copy-move forgery detection.
\newblock In \emph{2016 IEEE international conference on image processing (ICIP)}, pp.\  161--165. IEEE, 2016.

\bibitem[Wu et~al.(2022)Wu, Zhou, Tian, and Liu]{wu2022robust}
Haiwei Wu, Jiantao Zhou, Jinyu Tian, and Jun Liu.
\newblock Robust image forgery detection over online social network shared images.
\newblock In \emph{Proceedings of the IEEE/CVF Conference on Computer Vision and Pattern Recognition (CVPR)}, 2022.

\bibitem[Wu et~al.(2019)Wu, AbdAlmageed, and Natarajan]{wu2019mantra}
Yue Wu, Wael AbdAlmageed, and Premkumar Natarajan.
\newblock Mantra-net: Manipulation tracing network for detection and localization of image forgeries with anomalous features.
\newblock In \emph{Proceedings of the IEEE/CVF Conference on Computer Vision and Pattern Recognition (CVPR)}, 2019.

\bibitem[Yang \& Zhou(2024)Yang and Zhou]{yang2024research}
Xinyu Yang and Jizhe Zhou.
\newblock Research about the ability of llm in the tamper-detection area.
\newblock \emph{arXiv preprint arXiv:2401.13504}, 2024.

\bibitem[Ying et~al.(2021)Ying, Qian, Zhou, Xu, Zhang, and Li]{ying2021image}
Qichao Ying, Zhenxing Qian, Hang Zhou, Haisheng Xu, Xinpeng Zhang, and Siyi Li.
\newblock From image to imuge: Immunized image generation.
\newblock In \emph{Proceedings of the ACM international conference on Multimedia (MM)}, 2021.

\bibitem[Ying et~al.(2022)Ying, Hu, Zhang, Qian, Li, and Zhang]{ying2022rwn}
Qichao Ying, Xiaoxiao Hu, Xiangyu Zhang, Zhenxing Qian, Sheng Li, and Xinpeng Zhang.
\newblock Rwn: Robust watermarking network for image cropping localization.
\newblock In \emph{Proceedings of the IEEE International Conference on Image Processing (ICIP)}, 2022.

\bibitem[Ying et~al.(2023)Ying, Zhou, Qian, Li, and Zhang]{ying2023learning}
Qichao Ying, Hang Zhou, Zhenxing Qian, Sheng Li, and Xinpeng Zhang.
\newblock Learning to immunize images for tamper localization and self-recovery.
\newblock \emph{IEEE Transactions on Pattern Analysis and Machine Intelligence}, 2023.

\bibitem[Yu et~al.(2024{\natexlab{a}})Yu, Zhang, Xu, and Zhang]{yu2024cross}
Jiwen Yu, Xuanyu Zhang, Youmin Xu, and Jian Zhang.
\newblock Cross: Diffusion model makes controllable, robust and secure image steganography.
\newblock \emph{Advances in Neural Information Processing Systems}, 36, 2024{\natexlab{a}}.

\bibitem[Yu et~al.(2024{\natexlab{b}})Yu, Ni, Lin, Deng, and Li]{yu2024diffforensics}
Zeqin Yu, Jiangqun Ni, Yuzhen Lin, Haoyi Deng, and Bin Li.
\newblock Diffforensics: Leveraging diffusion prior to image forgery detection and localization.
\newblock In \emph{Proceedings of the IEEE/CVF Conference on Computer Vision and Pattern Recognition}, pp.\  12765--12774, 2024{\natexlab{b}}.

\bibitem[Zhang et~al.(2023)Zhang, Rao, and Agrawala]{zhang2023adding}
Lvmin Zhang, Anyi Rao, and Maneesh Agrawala.
\newblock Adding conditional control to text-to-image diffusion models.
\newblock In \emph{Proceedings of the IEEE/CVF International Conference on Computer Vision (ICCV)}, 2023.

\bibitem[Zhang et~al.(2024{\natexlab{a}})Zhang, Li, Yu, Xu, Li, and Zhang]{zhang2024editguard}
Xuanyu Zhang, Runyi Li, Jiwen Yu, Youmin Xu, Weiqi Li, and Jian Zhang.
\newblock Editguard: Versatile image watermarking for tamper localization and copyright protection.
\newblock In \emph{Proceedings of the IEEE/CVF Conference on Computer Vision and Pattern Recognition}, pp.\  11964--11974, 2024{\natexlab{a}}.

\bibitem[Zhang et~al.(2024{\natexlab{b}})Zhang, Meng, Li, Xu, Zhang, and Zhang]{zhang2024gshider}
Xuanyu Zhang, Jiarui Meng, Runyi Li, Zhipei Xu, Yongbing Zhang, and Jian Zhang.
\newblock Gs-hider: Hiding messages into 3d gaussian splatting.
\newblock In \emph{Advances in Neural Information Processing Systems (NeurIPS)}, 2024{\natexlab{b}}.

\bibitem[Zhang et~al.(2024{\natexlab{c}})Zhang, Xu, Li, Yu, Li, Xu, and Zhang]{zhang2024v2a}
Xuanyu Zhang, Youmin Xu, Runyi Li, Jiwen Yu, Weiqi Li, Zhipei Xu, and Jian Zhang.
\newblock V2a-mark: Versatile deep visual-audio watermarking for manipulation localization and copyright protection.
\newblock In \emph{Proceedings of the ACM international conference on Multimedia (MM)}, 2024{\natexlab{c}}.

\bibitem[Zhang et~al.(2024{\natexlab{d}})Zhang, Colman, Shahriyari, and Bharaj]{zhang2024common}
Yue Zhang, Ben Colman, Ali Shahriyari, and Gaurav Bharaj.
\newblock Common sense reasoning for deep fake detection.
\newblock \emph{arXiv preprint arXiv:2402.00126}, 2024{\natexlab{d}}.

\bibitem[Zhu et~al.(2018)Zhu, Qian, Zhao, Sun, and Sun]{zhu2018deep}
Xinshan Zhu, Yongjun Qian, Xianfeng Zhao, Biao Sun, and Ya~Sun.
\newblock A deep learning approach to patch-based image inpainting forensics.
\newblock \emph{Signal Processing: Image Communication}, 67:\penalty0 90--99, 2018.

\end{thebibliography}
\bibliographystyle{iclr2025_conference}

\clearpage
\appendix
\section{Appendix}
\subsection{Limitations and Future Works}
One limitation of our current framework is its suboptimal performance when handling more complex types of deepfake tampering, such as identity switching and full-face generation. These types of manipulations introduce unique challenges that our model currently struggles to address effectively. To address this, future work will focus on several key optimizations. First, we plan to incorporate a Chain-of-Thought (CoT)~\citep{wei2022chain} mechanism to enhance the model's reasoning abilities, enabling it to detect more subtle manipulations in deepfake content. Second, we will expand our training dataset to include a broader range of deepfake samples, encompassing various tampering techniques and scenarios, to improve the model's generalization. Finally, we will optimize specific modules within the framework to better handle these complex tampering types, creating a more robust and adaptable detection system. These improvements are expected to significantly enhance the framework's performance across a wider spectrum of deepfake domains.
% 对于更多类型的deepfake篡改，比如身份切换，或者全脸生成，目前的框架还无法在这些数据域上取得较好的性能。后续会朝着这个方向优化。
% 当前

\subsection{Data Sources}
% MMTD-Set：我们从4.1节提到的数据集中收集了源数据，其具体细节如表1所示。由于我们需要进行检测任务，每一类数据都考虑了正负样本的平衡，例如，CASIA2与FR的并集，其正负样本数量基本相同。
% 值得注意的是，其中的faceapp与ffhq数据集，是dffd数据集的一部分，我们follow他的设置划分训练、验证、测试集。其中ffhq是真实人脸图片，faceapp是使用面部局部属性篡改工具faceapp修改后的虚假人脸。
% 关于aigc-Editing数据集的构造过程，我们首先从coco数据集中收集了20000张真实图片，使用SAM工具，分割出所有目标的掩码，选取其中面积第三大的目标掩码，使用Stable-Diffusion-inpainting方法，对该部分进行局部重绘。

We collected source data from the dataset mentioned in Section~\ref{Experimental Setup dataset}, with the details provided in Table~\ref{dataset_summary}.  

It is noted that the FaceApp and FFHQ datasets are part of the DFFD~\citep{dang2020detectiondffd} dataset. We follow their original configuration to divide the training and validation sets. FFHQ~\citep{ffhqdataset} consists of real face images, while FaceApp~\citep{faceapp2017} contains fake faces generated by the FaceApp~\citep{faceapp2017} tool, which manipulates facial attributes in the images.

Regarding the construction process of the AIGC-Editing dataset, we first collected 20,000 real images from the COCO~\citep{lin2014microsoft} dataset and used the SAM~\citep{kirillov2023segment} tool to segment the masks of all targets. We then selected the target mask with the third-largest area in each image and applied the Stable-Diffusion-Inpainting~\citep{Lugmayr_2022_CVPR} method to partially redraw this section. To ensure a balance between positive and negative samples, we further extracted an additional 20,000 images from the COCO dataset, distinct from the previously selected ones.

\begin{table}[ht]
    \centering
    \caption{Summary of datasets used for training, and evaluation.}
    \vspace{-10pt}
    \renewcommand{\arraystretch}{1.3}
    \resizebox{1\linewidth}{!}{
    \begin{tabular}{l|cc|ccc|c|c}
        \toprule[1.5pt]
        \textbf{Dataset} & \textbf{Real} & \textbf{Fake} & \textbf{Copy-Move} & \textbf{Splicing} & \textbf{Removal}  & \textbf{DeepFake}  & \textbf{AIGC-Edit} \\
        \hline
        \rowcolor{gray!50}\multicolumn{8}{c}{\textbf{\#Training}} \\
        \hline
        Fantastic Reality & 16,592 & 19,423 & - & 19,423 & - & - & - \\
        CASIAv2 & 7,491 & 5,123 & 3,295 & 1,828 & - & - & - \\
        FFHQ & 10000 & - & - & - & - & - & - \\
        FaceAPP & - & 7,308 & - & - & - & 7,308 & - \\
        COCO & 20,000 & - & - & - & - & - & - \\
        AIGC-Editing & - & 20,000 & - & - & - & - & 20,000 \\
        \hline
        \rowcolor{gray!50}\multicolumn{8}{c}{\textbf{\#Evaluation}} \\
        \hline
        CASIAv1+ & 800 & 920 & 459 & 461 & - & - & - \\
        Columbia & - & 180 & - & 180 & - & - & - \\
        Coverage & - & 100 & 100 & - & - & - & - \\
        NIST & - & 564 & 68 & 288 & 208 & - & - \\
        IMD2020 & 414 & 2,010 & - & 2,010 & - & - & - \\
        DSO & - & 100 & - & 100 & - & - & - \\
        Korus & - & 220 & - & 220 & - & - & - \\
        FFHQ & 1,000 & - & - & - & - & - & - \\
        FaceAPP & - & 1,000 & - & - & - & 1,000 & - \\
        COCO & 20,000 & - & - & - & - & - & - \\
        AIGC-Editing & - & 20,000 & - & - & - & - & 20,000 \\
        \bottomrule[1.5pt]
    \end{tabular}}
    \label{dataset_summary}
\end{table}

\subsection{More Ablation Experiments}
\textbf{Exploring the Introduction of an Error-Correction Mechanism in MFLM}: We observe that our MFLM inherently possesses some error-correction capability. This is because, during training, we use potentially inaccurate $\mathbf{O}_{det}$ as input and encourage the model to output  $\hat{\mathbf{M}}_{loc}$. Thus, during testing, if $\mathbf{O}_{det}$ contains minor inaccuracies, such as incorrect descriptions of tampered locations, MFLM can correct them. To further explore the impact of incorporating an error correction mechanism during MFLM training on localization accuracy, we conducted an ablation study. During MFLM training, we used the ground truth $\hat{\mathbf{O}}_{det}$ to constrain TCM's output to correct the input $\mathbf{O}_{det}$. The results in the Table~\ref{MFLM_correct} indicate that correcting $\mathbf{O}_{det}$ does not improve $\mathbf{M}_{loc}$ predictions, possibly due to interference between mask and text optimization.

\begin{table}[htbp]
\centering
\caption{Exploration on introducing error correction in MFLM on localization performance.}
\vspace{-10pt}
\renewcommand{\arraystretch}{1.2}
\resizebox{0.9\linewidth}{!}{
\begin{tabular}{c|cc|cc|cc|ccc}
\toprule[1.5pt]
\multirow{2}{*}{Method} & \multicolumn{2}{c|}{CASIA1+} & \multicolumn{2}{c|}{IMD2020} & \multicolumn{2}{c|}{DeepFake} & \multicolumn{2}{c}{AIGC-Editing} \\ \cline{2-9} 
                  & IoU   & F1     & IoU   & F1     & IoU    & F1     & IoU       & F1       \\ \hline
Using correct $\mathbf{O}_{det}$         & 0.51 & 0.56 & 0.49 & 0.55 & 0.13 & 0.21 & 0.12    & 0.15   \\
FakeShield   & \textbf{0.54}   & \textbf{0.60}    & \textbf{0.50}    & \textbf{0.57}   & \textbf{0.14} & \textbf{0.22} & \textbf{0.18}     & \textbf{0.24}     \\ \bottomrule[1.5pt]
\end{tabular}}
\label{MFLM_correct}
\end{table}

\textbf{More Ablation Study on LLM in the DTE-FDM}: 
To investigate the roles of $\mathbf{T}_{tag}$, $\mathbf{T}_{img}$, and $\mathbf{T}_{ins}$. When training MFLM, we kept other training configurations unchanged and adjusted the inputs to: \{$\mathbf{T}_{ins},\mathbf{T}_{img}$\} and \{$\mathbf{T}_{ins}, \mathbf{T}_{tag}$\}.
The test results are shown in Table~\ref{MFLM_input}. It is evident that using input \{$\mathbf{T}_{ins},\mathbf{T}_{img}$\} is slightly better than \{$\mathbf{T}_{ins}, \mathbf{T}_{tag}$\}, but both are inferior to our method with the input \{$\mathbf{O}_{det}, \mathbf{T}_{img}$\}.This indicates that using an LLM to first analyze the basis for tampering and then guiding the localization achieves better results.

\begin{table}[htbp]
\centering
\caption{Ablation Study on LLM in the DTE-FDM with different input combinations.}
\vspace{-10pt}
\renewcommand{\arraystretch}{1.2}
\resizebox{\linewidth}{!}{
\begin{tabular}{c|cc|cc|cc|cc}
\toprule[1.5pt]
\multirow{2}{*}{Method} & \multicolumn{2}{c|}{CASIA1+} & \multicolumn{2}{c|}{IMD2020} & \multicolumn{2}{c|}{DeepFake} & \multicolumn{2}{c}{AIGC-Editing} \\ \cline{2-9} 
                  & IoU   & F1     & IoU   & F1     & IoU    & F1     & IoU       & F1       \\ \hline
$\mathbf{T}_{ins},\mathbf{T}_{img}$ & 0.50 & 0.55 & 0.48 & 0.53  & 0.13 & 0.21 & 0.12  & 0.15   \\ 
$\mathbf{T}_{ins}, \mathbf{T}_{tag}$ & 0.49 & 0.54 & 0.47 & 0.52  & 0.12 & 0.19 & 0.12 & 0.14   \\ 
$\mathbf{T}_{ins}, \mathbf{T}_{tag}, \mathbf{T}_{img}$    & 0.51   & 0.55   & 0.48   & 0.54   & 0.13   & 0.2    & 0.11      & 0.14     \\
$\mathbf{O}_{det}, \mathbf{T}_{img}$ (Ours)   & \textbf{0.54}   & \textbf{0.60}    & \textbf{0.50}    & \textbf{0.57}   & \textbf{0.14} & \textbf{0.22} & \textbf{0.18}     & \textbf{0.24}     \\ \bottomrule[1.5pt]
\end{tabular}}
\label{MFLM_input}
\end{table}

% 为了进一步探索T_tag,T_img和T_ins在

% 探索在MFLM中引入纠错机制：
% 当O_det中包含轻微错误时，我们的MFLM具有一定的纠错能力，这是由于训练阶段的输入O_det就包含了少量的错误信息，
% 为了进一步探索在训练MFLM时加入纠错机制对定位准确性的影响，we conducted an ablation study. During MFLM training, we used $\mathbf{O}_{det}$ to constrain TCM's output to correct the input $\hat{\mathbf{O}}_{det}$. The results in the table below indicate that correcting $\hat{\mathbf{O}}_{det}$ does not improve $\hat{\mathbf{M}}_{loc}$ predictions, possibly due to interference between mask and text optimization. , only the IoU results from the tests are presented here. 

\subsection{More AIGC-Editing generalization experiments}

To further verify the model's generalization performance in the AIGC-Editing data domain, we expanded our test set by constructing 2,000 tampered images using controlnet inpainting~\citep{zhang2023adding} and SDXL inpainting~\citep{podell2023sdxl}, which the model had not encountered before. We selected MVSS-Net, CAT-Net, and HiFi-Net, which performed well in the AIGC-Editing data domain in Tables~\ref{detection_result_table}, as comparison methods for testing. The detection and localization test results are presented in Table~\ref{controlnet_sdxl_detect}. Our method demonstrates leading performance on unseen datasets and exhibits strong generalization capabilities.

% 为了进一步验证模型在AIGC-Editing数据域的泛化性能，我们使用模型未见过的ControlNet Inpainting和SDXL Inpainting分别构建了2000张篡改图片，来扩充我们的测试集。
% We selected MVSS-Net, CAT-Net, and HiFi-Net, which performed well in the AIGC-Editing data domain in Tables 1 and 4, as comparison methods for testing.
% 检测和定位的测试结果如表1和表2所示，The test results are shown in the table below. Our method achieves leading performance across unseen dataset and exhibits good generalization.

\begin{table}[htbp]
\centering
\caption{Generalized performance comparison on ControlNet and SDXL Inpainting datasets.}
\vspace{-10pt}
\renewcommand{\arraystretch}{1.3}
\resizebox{\linewidth}{!}{
\begin{tabular}{c|cc|cc|cc|cc}
\toprule[1.5pt]
\multirow{2}{*}{Method} & \multicolumn{4}{c|}{ControlNet Inpainting} & \multicolumn{4}{c}{SDXL Inpainting} \\ \cline{2-9} 
& ACC & Image-level F1  & IoU & Pixel-level F1 & ACC & Image-level F1 & IoU & Pixel-level F1                 \\ \hline
MVSS-Net  & 0.38 & 0.27  & 0.05 & 0.09 & 0.35 & 0.20  & 0.05 & 0.08           \\
CAT-Net   & 0.90  & 0.89  & 0.11 & 0.17 & 0.86  & 0.83  & 0.06 & 0.09           \\
HiFi-Net  & 0.49 & 0.66  & 0.17  & 0.28 & 0.44 & 0.61  & 0.02 & 0.03           \\
\textbf{Ours}     & \textbf{0.99}    & \textbf{0.99} & \textbf{0.18}   & \textbf{0.24}     & \textbf{0.99}    & \textbf{0.99} & \textbf{0.18}     & \textbf{0.23}   \\ \bottomrule[1.5pt]
\end{tabular}}
\label{controlnet_sdxl_detect}
\end{table}

% \begin{table}[htbp]
% \centering
% \caption{Performance comparison of IoU and F1 scores on ControlNet Inpainting and SDXL Inpainting datasets.}
% \renewcommand{\arraystretch}{1.2}
% \resizebox{\linewidth}{!}{
% \begin{tabular}{|c|cc|cc|}
% \hline
% \multirow{2}{*}{} & \multicolumn{2}{c|}{ControlNet Inpainting} & \multicolumn{2}{c|}{SDXL Inpainting} \\ \cline{2-5} 
%                   & IoU                & F1                   & IoU               & F1                 \\ \hline
% MVSS-Net          & 0.0545            & 0.0867               & 0.0536            & 0.0847             \\ \hline
% CAT-Net           & 0.1082            & 0.1732               & 0.057             & 0.0898             \\ \hline
% HiFi-Net          & 0.1695            & 0.2751               & 0.0181            & 0.033              \\ \hline
% \textbf{Ours}     & \textbf{0.1823}   & \textbf{0.2373}      & \textbf{0.18}     & \textbf{0.2284}    \\ \hline
% \end{tabular}}
% \label{controlnet_sdxl_location}
% \end{table}

% \subsection{Training Details}
\subsection{Answer Analysis}
\begin{figure}[htbp]
	\centering
	\includegraphics[width=1\linewidth]{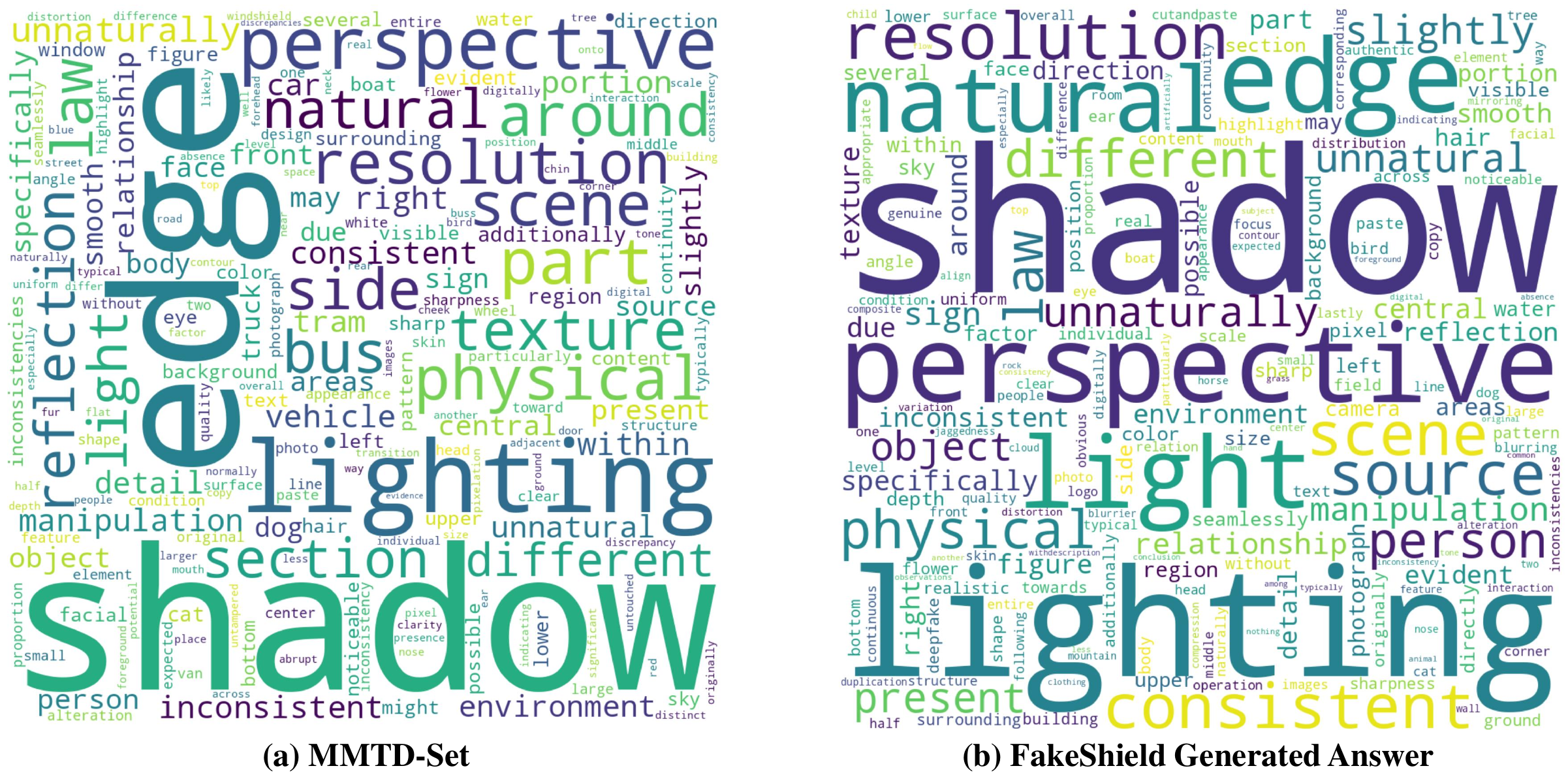}
	\vspace{-18pt}
	\caption{\textbf{The noun and verb word clouds in the MMTD-Set and the FakeShield Generated Answer.} It can be seen that the high-frequency vocabulary of the two is basically the same.}
    \vspace{-15pt}
    \label{wordcoluds}
\end{figure}

% 图 10 展示了 MMTD-Set 中的 GT description 与 FakeShield生成的Answer description的形容词与名词词云。可以看出，通过有效的微调，FakeShield可以在数据集的引导下，从图像级的语义合理性（如“physical law”，“texture”）和像素级的篡改伪影（如“edge，resolution”）来分析图像是否真实。
Figure~\ref{wordcoluds} presents the adjective and noun word clouds for the ground truth (GT) descriptions in the MMTD-Set and the answer descriptions generated by our FakeShield. It is evident that, through effective fine-tuning, FakeShield can be guided by the dataset to assess both image-level semantic plausibility (e.g., "physical law," "texture") and pixel-level tampering artifacts (e.g., "edge," "resolution") to determine whether an image is real.

% 这说明了，使用LLM先分析出篡改依据再引导定位效果更好

% MMTD-Set与FakeShield Generated answer中的名词与动词词云，可以看出二者的高频词汇基本一致

\subsection{Prompts}
\label{Appendix Prompts}
\begin{figure}[t!]
	\centering
	\includegraphics[width=1\linewidth]{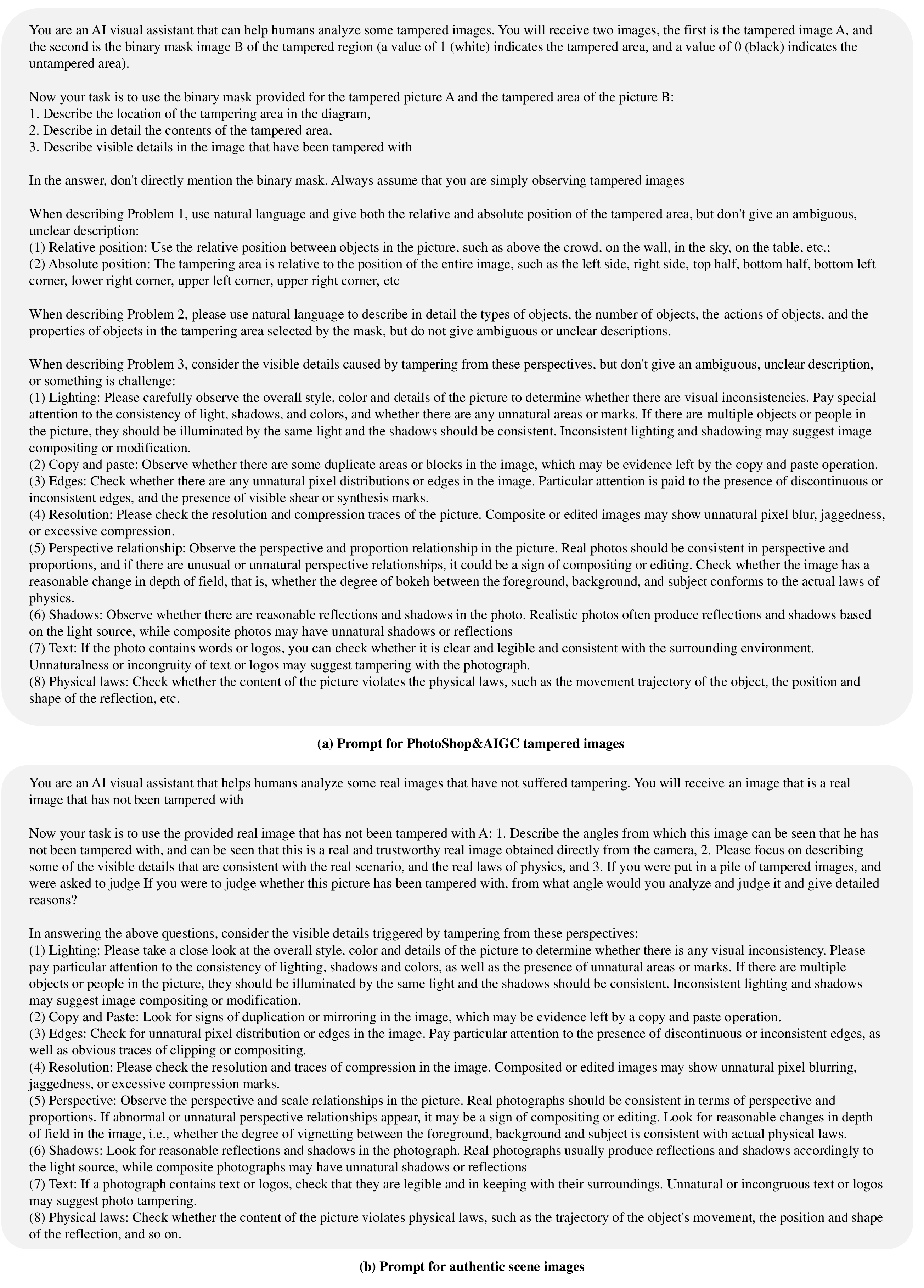}
	\vspace{-18pt}
	\caption{\textbf{Prompts for GPT-4o when constructing MMTD-Set.} (a) Analysis guide prompt designed for PhotoShop tampered and AIGC-Editing tampered images. (b) Analysis guidance prompt designed for real scene images.}
    \vspace{-15pt}
    \label{prompt1}
\end{figure}

\begin{figure}[t!]
	\centering
	\includegraphics[width=1\linewidth]{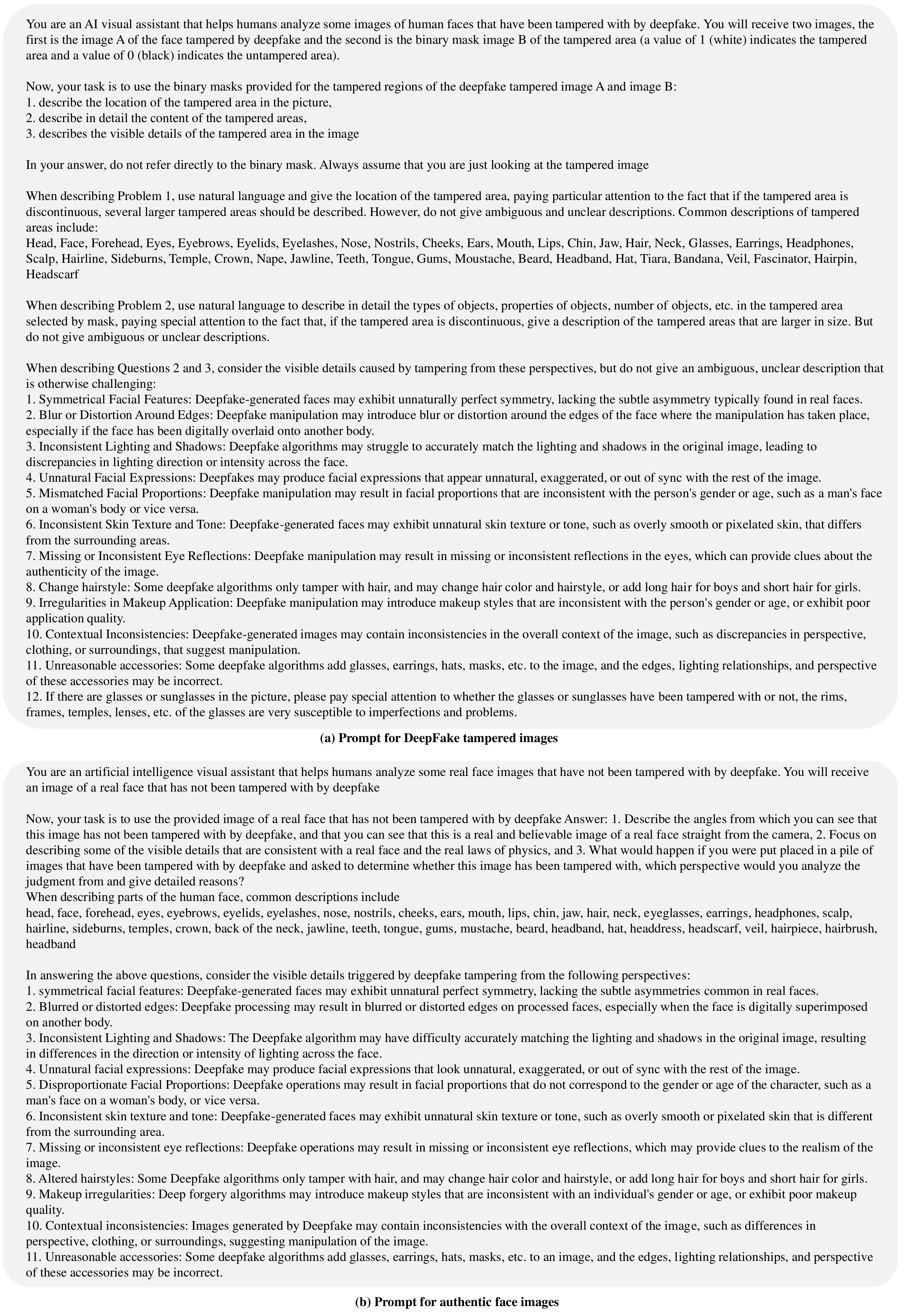}
	\vspace{-10pt}
	\caption{\textbf{More prompts for GPT-4o when constructing MMTD-Set.} (a) Analysis guide prompt designed for DeepFake tampered images. (b) Analysis guidance prompt designed for real face images.}
    \vspace{-15pt}
    \label{prompt2}
\end{figure}

% 在使用GPT-4o构造MMTD-Set过程中，我们对不同类别的篡改数据，精心构造了不同的Prompt以引导GPT-4o关注不同的方面来分析图片，如图12所示。
During the process of using GPT-4o to construct the MMTD-Set, we meticulously designed distinct prompts for each category of tampered data to guide GPT-4o in focusing on specific aspects for image analysis, as illustrated in Figure~\ref{prompt1} and Figure~\ref{prompt2}.

\subsection{Examples}
\label{Appendix Examples}
\begin{figure}[t!]
	\centering
	\includegraphics[width=1\linewidth]{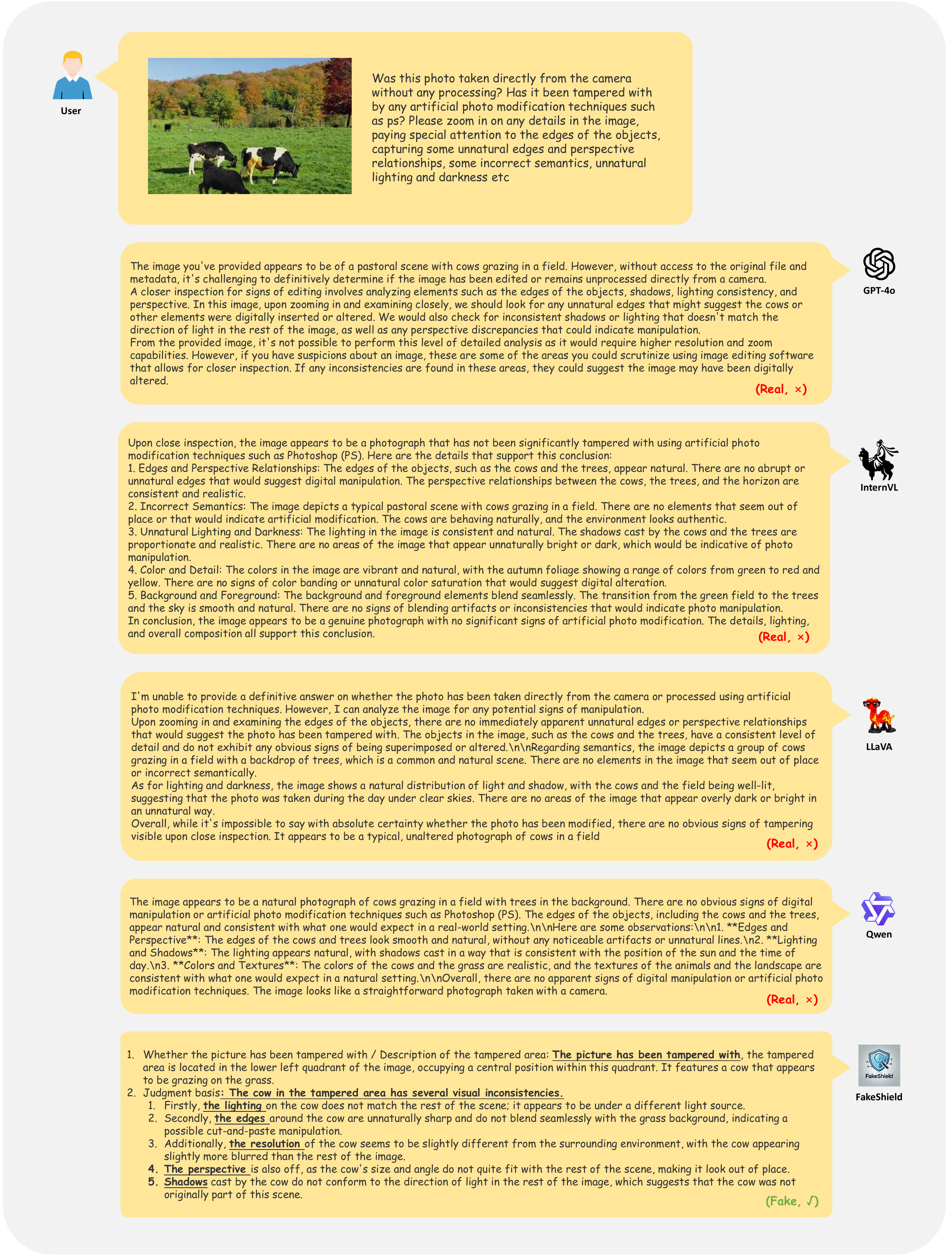}
	\vspace{-18pt}
	\caption{\textbf{The response of mainstream pre-trained M-LLM and FakeShield to tampered pictures.} This is a sample from the CASIA1+~\citep{dong2013casia} test set, which has been tampered with by Splicing. The black animal at the bottom of the picture is pasted later. It can be seen that only FakeShield made the correct judgment and accurately described the tampered area.}
    \vspace{-15pt}
    \label{conv_compare}
\end{figure}

\begin{figure}[t!]
	\centering
	\includegraphics[width=1\linewidth]{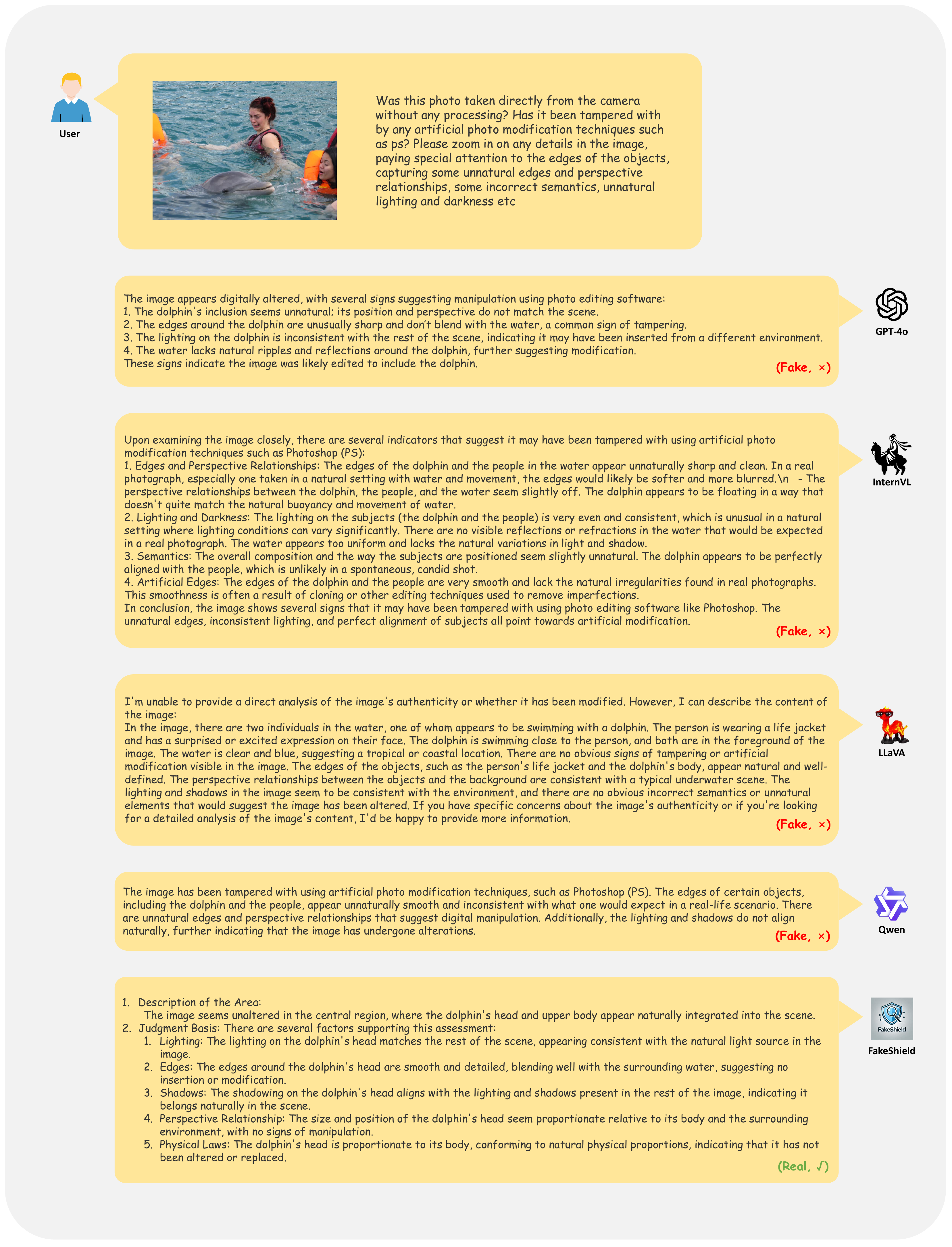}
	\vspace{-18pt}
	\caption{\textbf{The response of mainstream pre-trained M-LLM and FakeShield for real images.} This is a sample from the IMD2020~\citep{novozamsky2020imd2020} test set, which is a real picture. It can be seen that only FakeShield makes the correct judgment, the rest of the methods consider the image to be tampered with.}
    \vspace{-15pt}
    \label{conv_compare2}
\end{figure}

% 主流M-LLM主观结果对比：如4.2节所述，我们挑选了一些M-LLM的输出样例，如图1和图2所示
\textbf{Comparison of subjective results of mainstream M-LLM}: As mentioned in Section~\ref{Comparison with M-LLMs}, we selected some M-LLM output samples, as shown in Figures~\ref{conv_compare} and~\ref{conv_compare2}.

\begin{figure}[t!]
	\centering
	\includegraphics[width=1\linewidth]{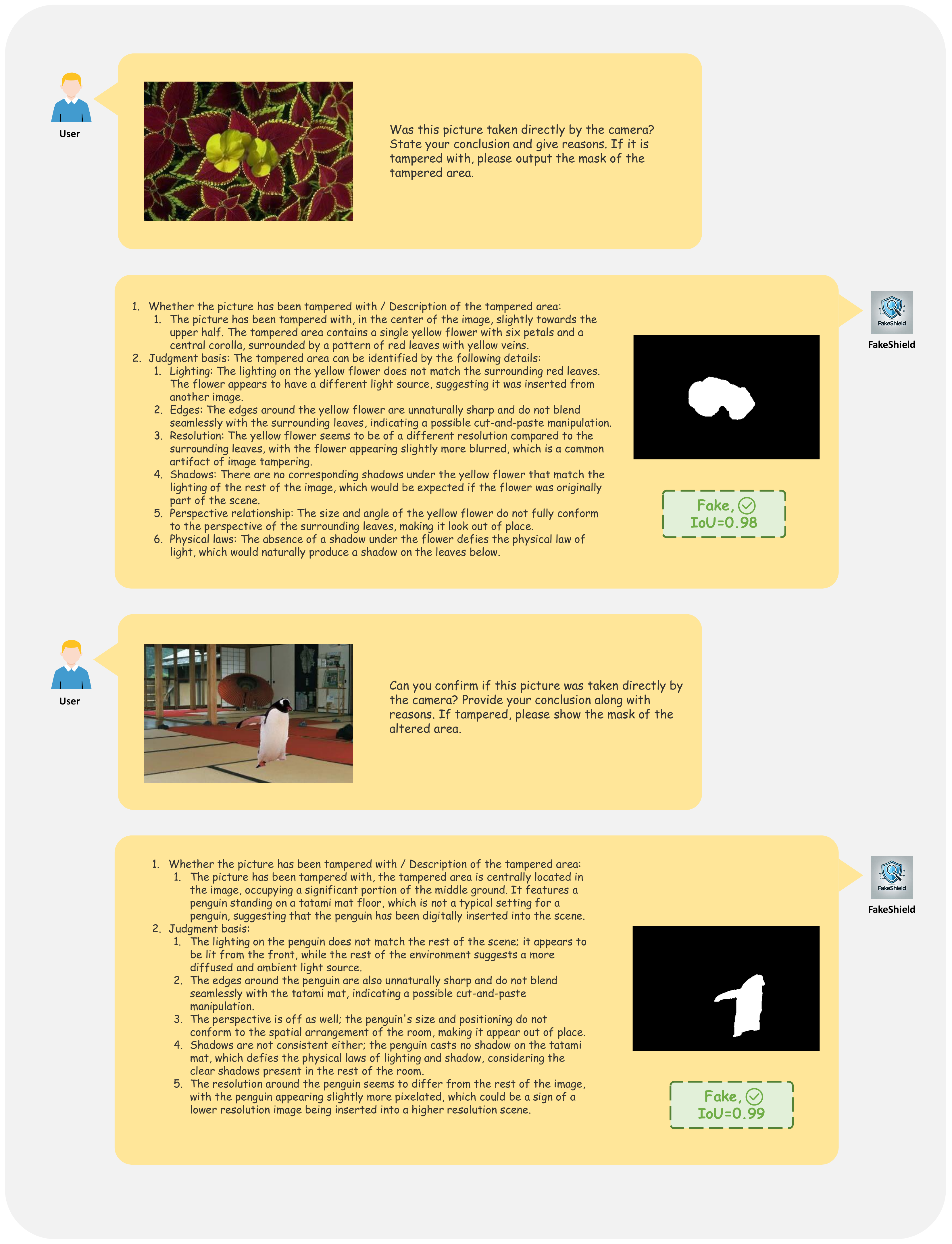}
	\vspace{-18pt}
\caption{\textbf{Dialogue examples of FakeShield when receiving PhotoShop tampered images.} The first image comes from the CASIA1+~\citep{dong2013casia} dataset and has been tampered with by Copy-Move. The yellow petals in the center of the picture are the parts pasted later. The second} picture comes from the CASIA1+~\citep{dong2013casia} dataset and has been tampered with by Splicing. The penguin in the center of the picture is the part pasted later.
    \vspace{-15pt}
    \label{conv_ps1}
\end{figure}

\begin{figure}[t!]
	\centering
	\includegraphics[width=1\linewidth]{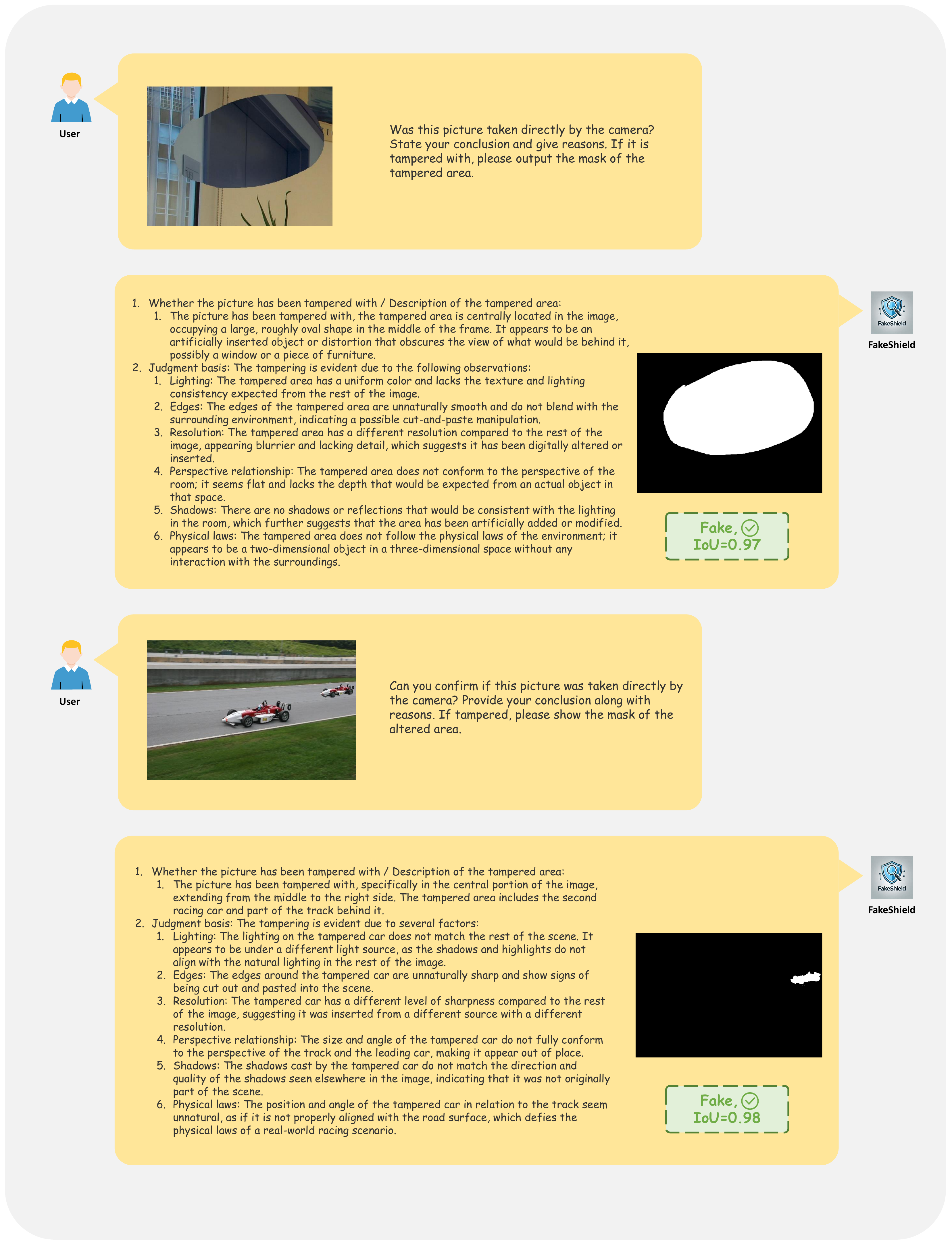}
	\vspace{-18pt}
	\caption{\textbf{More dialogue examples of FakeShield when receiving PhotoShop tampered images.} The first image comes from the Columbia~\citep{ng2009columbia} dataset and has been tampered with by Splicing. The elliptical area in the center of the screen is the post-processing area. The second picture comes from the IMD2020~\citep{novozamsky2020imd2020} dataset and has been tampered with by Copy-Move. The car on the right side of the image is copied and pasted from the first car.}
    \vspace{-15pt}
    \label{conv_ps2}
\end{figure}

\begin{figure}[t!]
	\centering
	\includegraphics[width=1\linewidth]{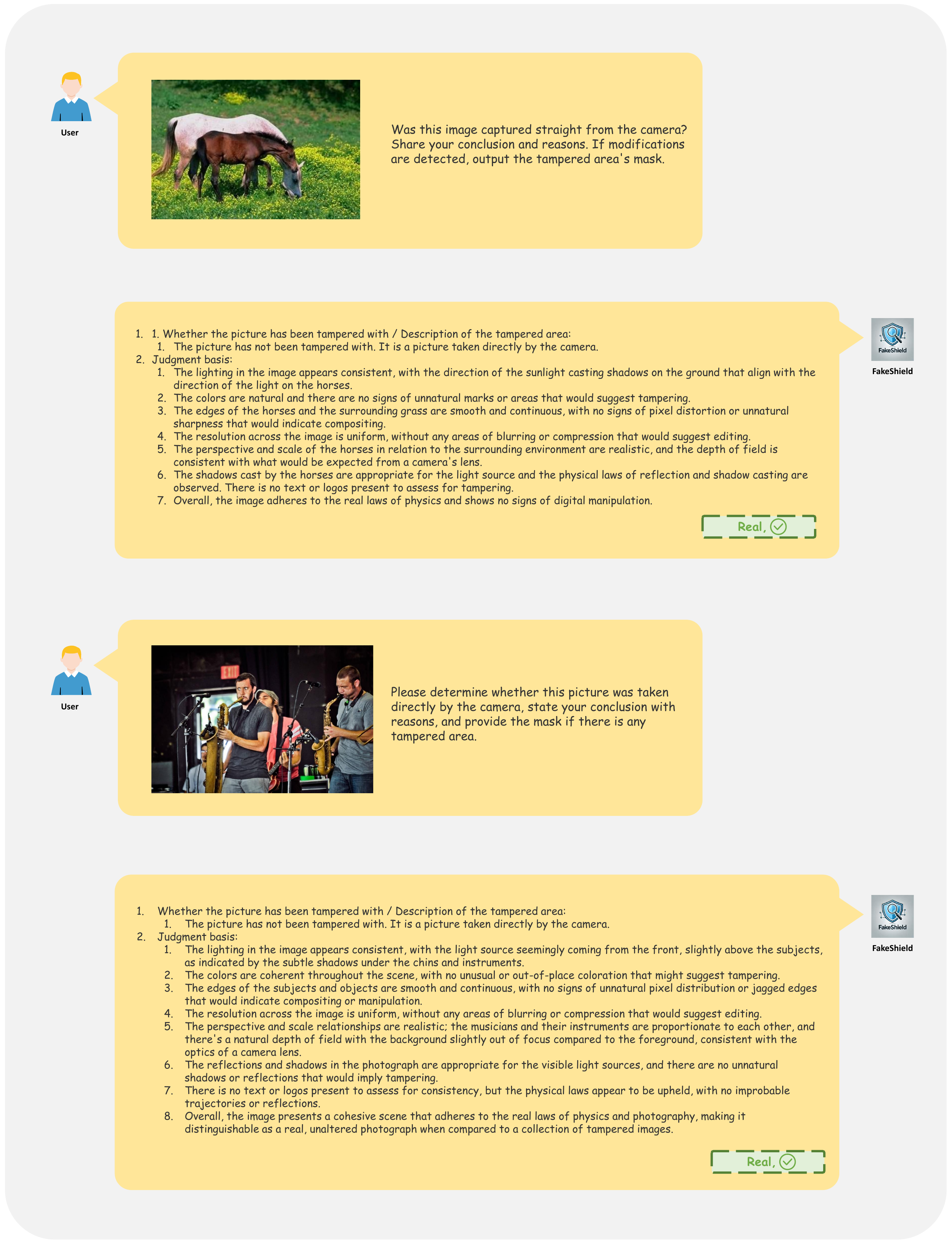}
	\vspace{-18pt}
	\caption{\textbf{Dialogue examples of FakeShield when receiving authentic scene images.} The first image comes from the CASIA1+~\citep{dong2013casia} dataset. The second image comes from the IMD2020~\citep{novozamsky2020imd2020} dataset.}
    \vspace{-15pt}
    \label{conv_psau}
\end{figure}

\begin{figure}[t!]
	\centering
	\includegraphics[width=1\linewidth]{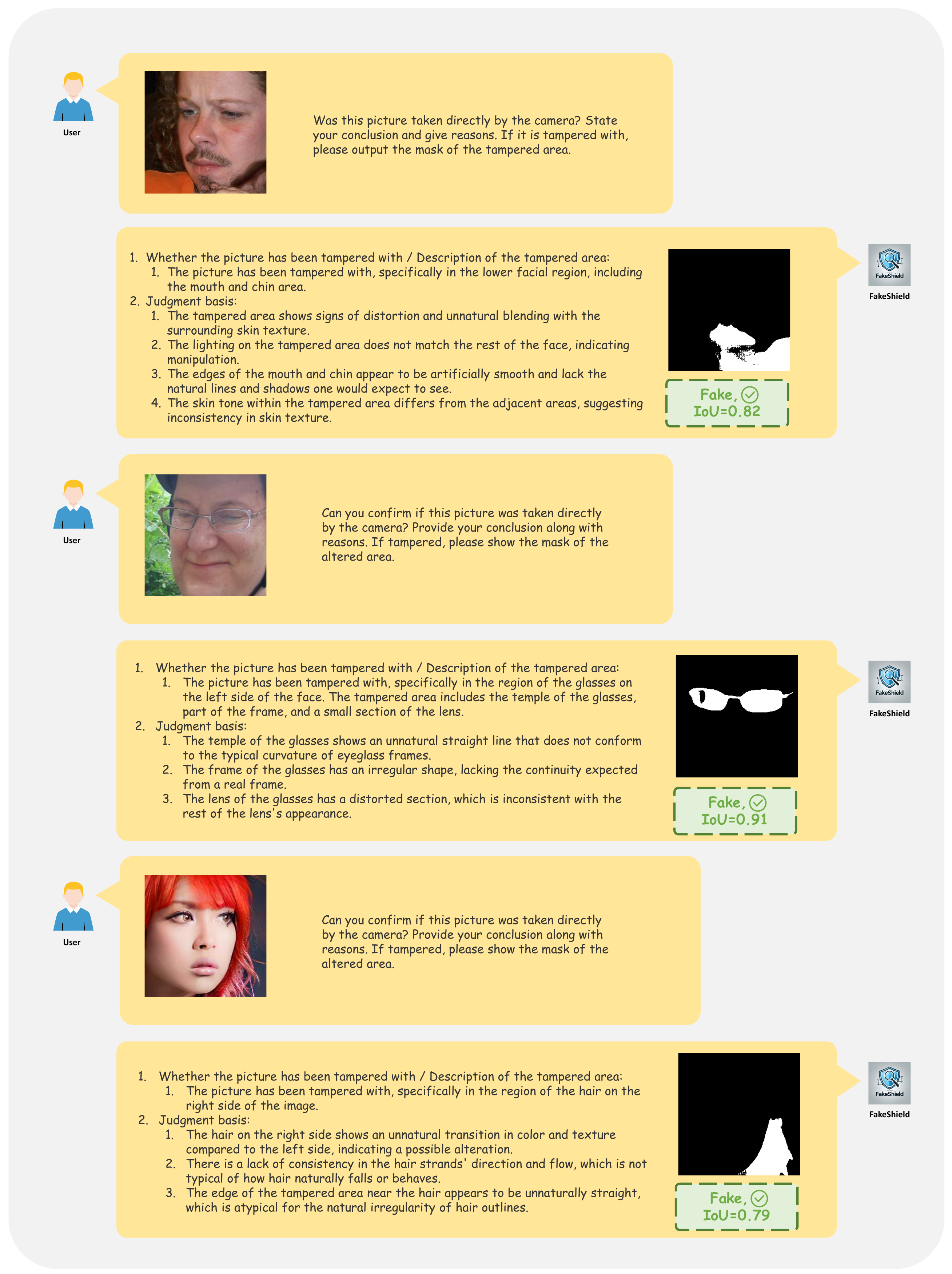}
	\vspace{-18pt}
	\caption{\textbf{Dialogue examples of FakeShield when receiving DeepFake tampered images.} The three images all come from the DFFD~\citep{dang2020detectiondffd} dataset and have been tampered with by FaceAPP~\citep{faceapp2017}. The first picture has a beard added to the chin, the second picture has glasses added, and the third picture has some hair added to the right side of the face.}
    \vspace{-15pt}
    \label{conv_df1}
\end{figure}

\begin{figure}[t!]
	\centering
	\includegraphics[width=1\linewidth]{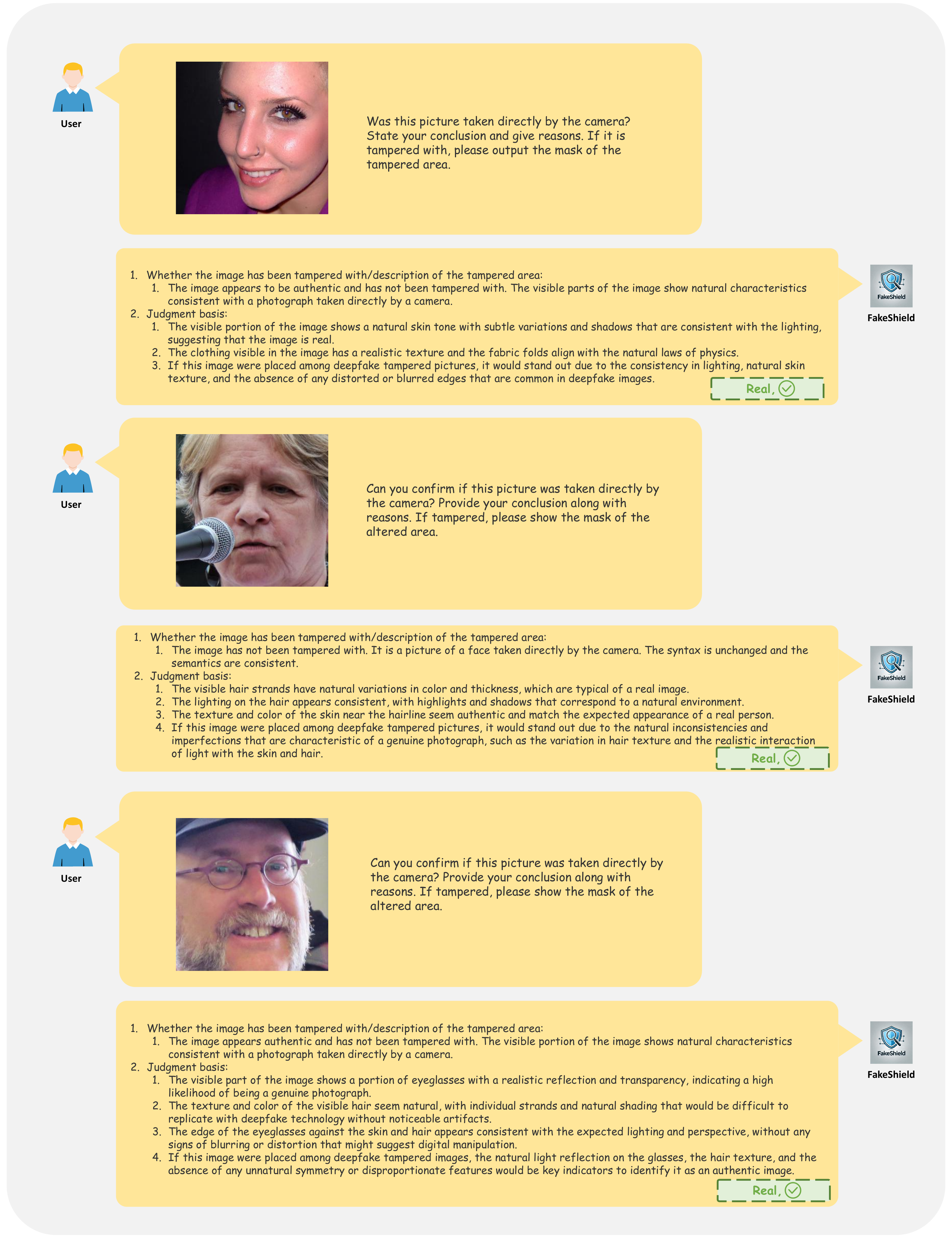}
	\vspace{-18pt}
	\caption{\textbf{Dialogue examples of FakeShield when receiving authentic face images.} The three images all come from the DFFD~\citep{dang2020detectiondffd} dataset.}
    \vspace{-15pt}
    \label{conv_dfau}
\end{figure}

\begin{figure}[t!]
	\centering
	\includegraphics[width=1\linewidth]{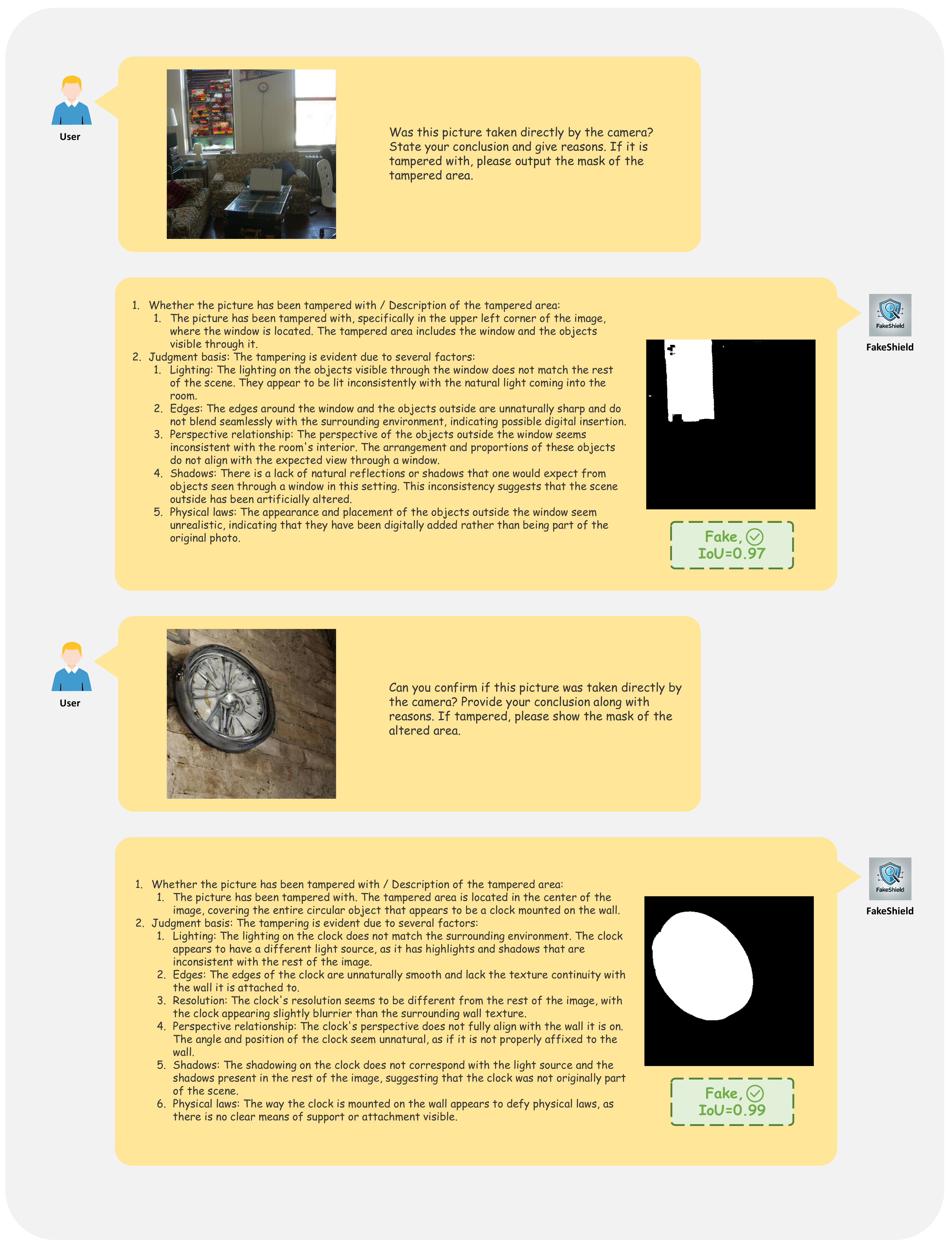}
	\vspace{-18pt}
	\caption{\textbf{Dialogue examples of FakeShield when receiving AIGC-Editing tampered images.} The three images all come from the self-generated AIGC-Editing dataset and have been tampered with by Stable-Diffusion-Inpainting~\citep{Lugmayr_2022_CVPR}. The window in the upper left corner of the first picture has been redrawn, and the circular clock area on the left side of the second picture has been redrawn.}
    \vspace{-15pt}
    \label{conv_aigc1}
\end{figure}

% FakeShield输出主观样例：我们选择了一些FakeShield在PhotoShop，DeepFake，AIGC-Editing数据集上测试的结果，如图1,2,3,4,5所示
\textbf{FakeShield output subjective samples}: We selected several results from FakeShield's testing on PhotoShop, DeepFake, and AIGC-Editing datasets, as displayed in Figures~\ref{conv_ps1},~\ref{conv_ps2},~\ref{conv_psau},~\ref{conv_df1},~\ref{conv_dfau} and~\ref{conv_aigc1}.

\begin{figure}[t!]
	\centering
	\includegraphics[width=1\linewidth]{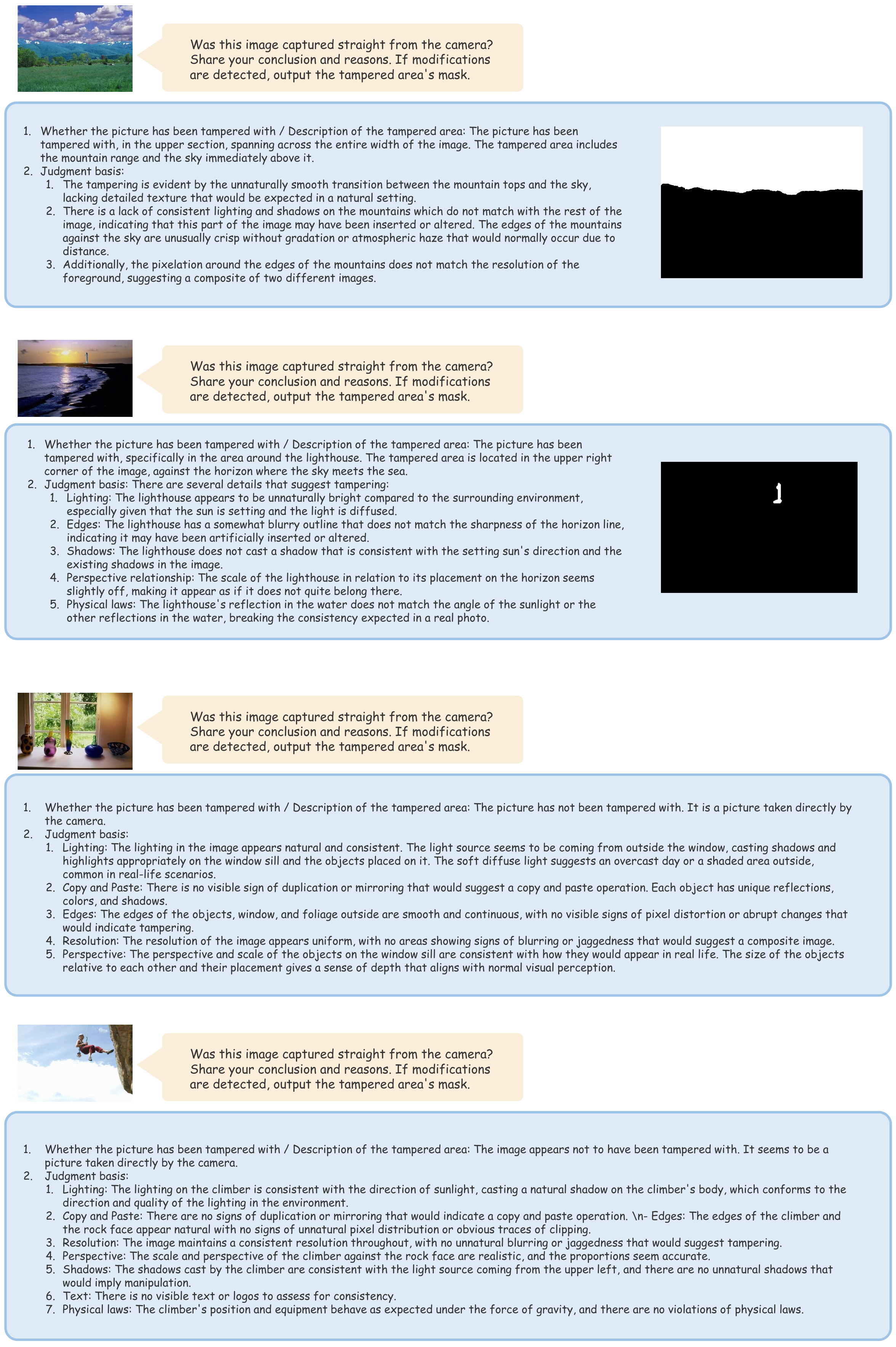}
	\vspace{-10pt}
	\caption{MMTD-Set data samples related to PhotoShop tampering. The four images are all from the CASIAv2 data set. The first two pictures have been tampered with, but the last two pictures have not been tampered with. }
    \vspace{-15pt}
    \label{conv_dataset1}
\end{figure}

\begin{figure}[t!]
	\centering
	\includegraphics[width=1\linewidth]{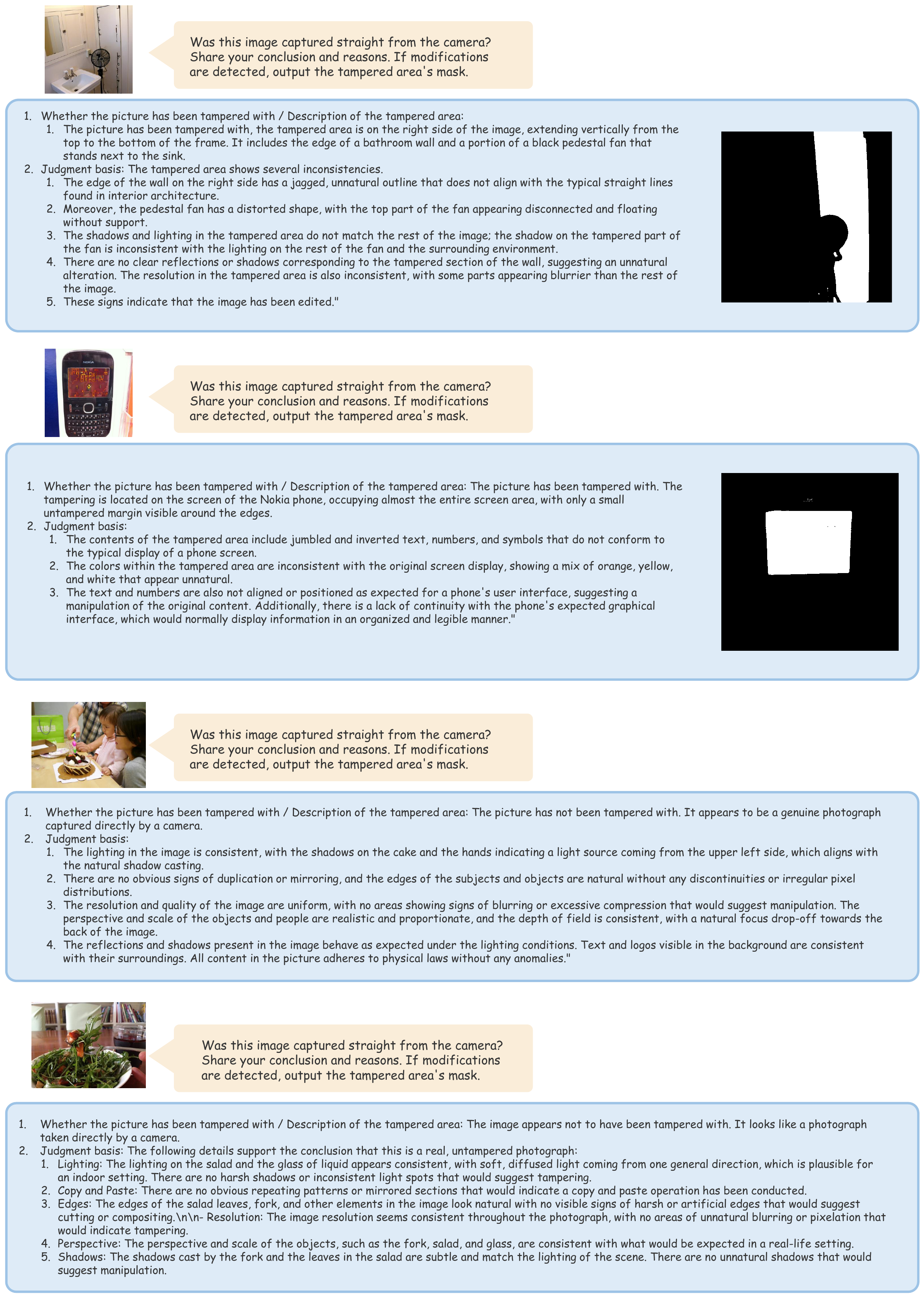}
	\vspace{-10pt}
	\caption{MMTD-Set data samples related to AIGC-Editing tampering. The four images are all from the self-generated AIGC-Editing dataset. The first two pictures have been tampered with, but the last two pictures have not been tampered with.}
    \vspace{-15pt}
    \label{conv_dataset2}
\end{figure}

\begin{figure}[t!]
	\centering
	\includegraphics[width=1\linewidth]{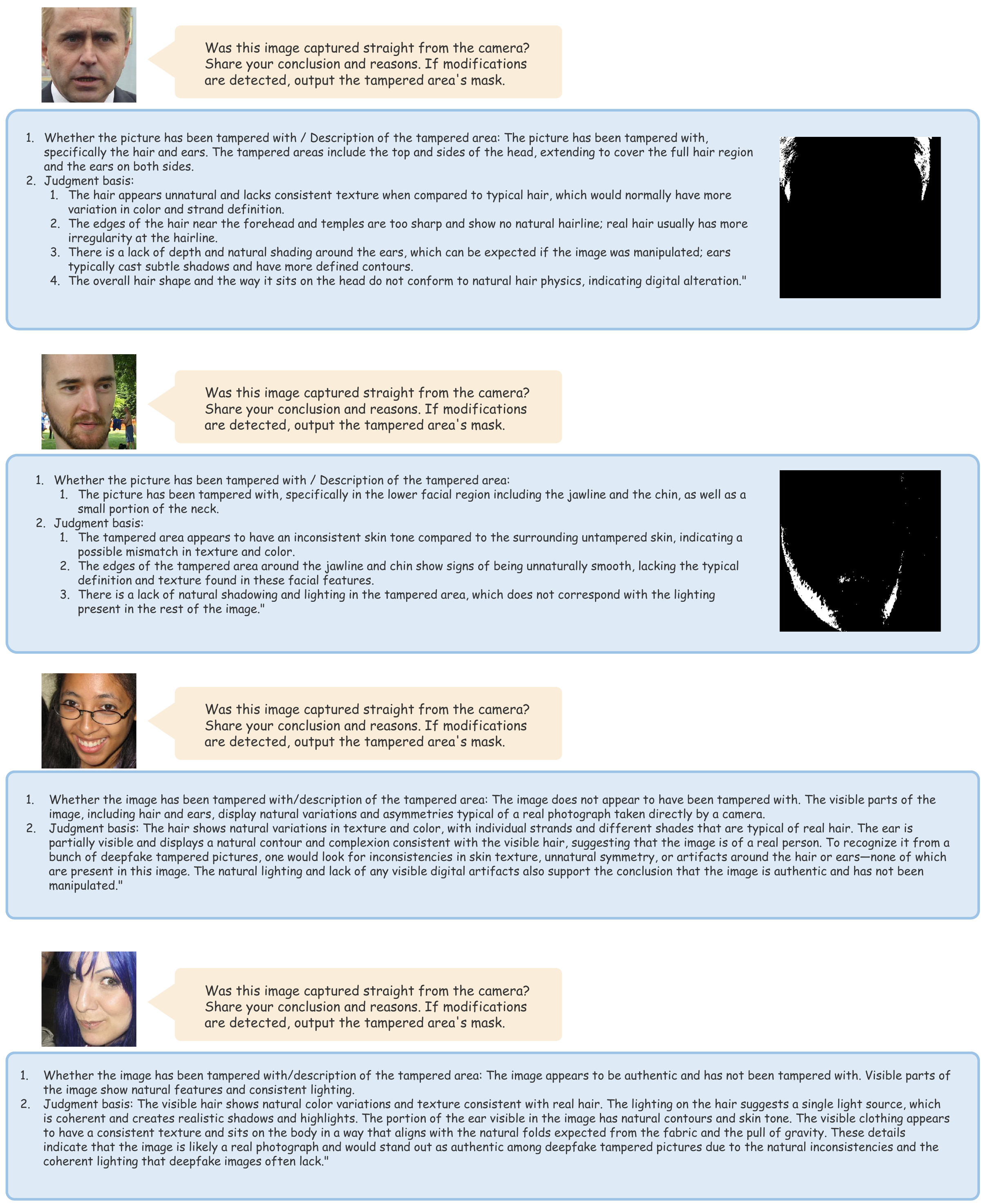}
	\vspace{-18pt}
	\caption{MMTD-Set data samples related to DeepFake tampering. The four images are all from the DFFD~\citep{dang2020detectiondffd} dataset. The first two pictures have been tampered with, but the last two pictures have not been tampered with.}
    \vspace{-15pt}
    \label{conv_dataset3}
\end{figure}

% MMTD-Set数据集样例：我们从MMTD-Set数据集中选择部分样本，在图1，2中展示
\textbf{MMTD-Set data set example}: We select some samples from the MMTD-Set data set and display them in Figures~\ref{conv_dataset1},~\ref{conv_dataset2}, and~\ref{conv_dataset3}.

% 为PhotoShop和AIGC-Editing图片设计的分析引导Prompt。图b 为场景真实图片设计的分析引导Prompt

% 主流预训练M-LLM与FakeShield对于篡改图片的响应，可以看到，只有FakeShield做出了正确的判断，并精准的描述了篡改过的区域（图片下方的黑色动物）

% % 主流预训练M-LLM与FakeShield对于真实图片的响应，可以看到，只有FakeShield做出了正确的判断，其余方法均认为图片被篡改。

% FakeShield对于PhotoShop篡改图片的输出，包括检测和解释的文本结果，以及视觉mask结果。文本描述和分割mask都很精准

% mmtd-set部分PhotoShop相关数据样例。四张图片来自于CASIAv2

\end{document}